# Central Dogma Transformer

Towards Mechanism-Oriented AI for Cellular Understanding

Nobuyuki Ota[1]

## Abstract

Understanding cellular mechanisms requires integrating information across DNA, RNA, and protein—the three molecular systems linked by the Central Dogma of molecular biology. While domain-specific foundation models have achieved success for each modality individually, they remain isolated, limiting our ability to model integrated cellular processes. Here we present the Central Dogma Transformer (CDT), an architecture that integrates pre-trained language models for DNA, RNA, and protein following the directional logic of the Central Dogma. CDT employs directional cross-attention mechanisms—DNA-to-RNA attention models transcriptional regulation, while RNA-to-Protein attention models translational relationships—producing a unified Virtual Cell Embedding that integrates all three modalities.

We validate CDT v1—a proof-of-concept implementation using fixed (non-cell-specific) RNA and protein embeddings—on CRISPRi enhancer perturbation data from K562 cells, achieving a Pearson correlation of 0.503, representing 63% of the theoretical ceiling set by cross-experiment variability (r = 0.797). Attention and gradient analyses provide complementary interpretive windows: in detailed case studies, these approaches highlight largely distinct genomic regions, with gradient analysis identifying a CTCF binding site that Hi-C data showed as physically contacting both enhancer and target gene. These results suggest that AI architectures aligned with biological information flow can achieve both predictive accuracy and mechanistic interpretability.

## 1. Introduction

Cellular function emerges from the coordinated action of three distinct molecular systems: DNA, which encodes genetic instructions in nucleotide sequences; RNA, which serves as the intermediary carrying genetic messages and regulating gene expression; and proteins, which execute cellular functions as the primary molecular machines. These three systems are linked by the Central Dogma of molecular biology [1], wherein information flows from DNA to RNA through transcription, and from RNA to protein through translation. This directional flow constitutes the fundamental organizing principle of cellular biology.

A central challenge in computational biology is developing AI systems that integrate knowledge across these three molecular modalities. Current AI systems, however, remain fundamentally opaque—their internal representations lack interpretable meaning. While such opacity may be acceptable in engineering applications where prediction accuracy alone suffices, science demands more. The goal of scientific research is not merely to predict phenomena but to understand the mechanisms that produce them. In biology, this means understanding how molecular interactions give rise to cellular behavior—insight that opaque models cannot provide.

[1]Independent Researcher, Burlingame, CA, USA. ORCID: 0009-0006-6570-9450

The Transformer architecture, introduced by Vaswani et al. [2], has been central to recent advances in AI. Its self-attention mechanism enables learning long-range dependencies in sequences, making it particularly suited for biological data where distant elements—such as enhancers and their target genes separated by tens to hundreds of kilobases—often interact functionally. This architectural innovation has enabled the development of powerful foundation models across biological domains.

At the DNA level, models such as Enformer [3], Nucleotide Transformer [4], HyenaDNA [5], Evo [6], and Evo 2 [7] have learned the grammar of genomic sequences, with Evo 2 extending context to one million base pairs. For RNA, scGPT [8] and Geneformer [9] learn gene-gene relationships from single-cell transcriptomic data. At the protein level, AlphaFold [10] revolutionized structure prediction, ESM-C [11] encodes structural and functional properties from amino acid sequences, and ProteomeLM [12] learns protein relationships within proteome context. Yet these models remain isolated from one another: knowledge learned by protein structure models does not inform genomic sequence models, and expression patterns learned from transcriptomic data do not guide DNA sequence interpretation. This isolation contradicts the integrated nature of cellular biology, where signals propagate continuously from DNA to RNA to protein.

Previous attempts at multi-modal integration have largely relied on concatenating embeddings from different modalities before processing through neural networks [13, 14, 15, 16]. While concatenation allows joint processing, it treats DNA, RNA, and protein as interchangeable inputs without inherent order, providing no mechanism for learning directional regulatory relationships. More recent approaches employ cross-attention for multi-modal integration [17], but still treat modalities symmetrically without encoding the directional logic of biological information transfer.

To address this limitation, we propose the Central Dogma Transformer (CDT), an architecture that integrates DNA, RNA, and protein language models following the directional logic of the Central Dogma. The core design principle is that computational structure should reflect biological structure. CDT implements information flow through two sequential cross-attention layers: in DNA-to-RNA attention, RNA representations query DNA positions to learn which genomic regions regulate each gene (mirroring transcription); in RNA-to-Protein attention, protein representations query RNA to capture translational relationships. This directional design produces interpretable attention weights that correspond to specific biological relationships. These representations converge into a Virtual Cell Embedding (VCE)—a unified vector representing cellular state. CDT is designed as a modular platform rather than a monolithic model, enabling component updates as improved foundation models become available.

The architecture supports two complementary approaches for biological interpretation [18]. Attention analysis reveals which molecular features the model considers during forward inference; gradient analysis traces prediction-relevant features backward through the computational graph to specific genomic positions [19]. Because CDT's architecture mirrors the biological cascade, gradient-based feature attribution may carry mechanistic meaning—though this interpretation requires experimental validation. Notably, attention and gradient analyses often highlight different genomic regions, suggesting these approaches capture complementary aspects of the learned representations.

We present CDT v1 as a proof of concept using CRISPRi enhancer perturbation data, with fixed (non-cell-specific) RNA and protein embeddings. CDT predicts enhancer perturbation effects from DNA sequence (Pearson r = 0.503, representing 63% of the theoretical ceiling set by cross-experiment variability). Cross-attention patterns are largely uniform across samples due to fixed scGPT gene token embeddings, while gradient analysis identifies specific genomic features critical for individual predictions—for example, a CTCF binding site [20] that Hi-C data [21] showed as physically contacting both the enhancer and the gene promoter. This finding suggests CDT learned features related to chromatin architecture from sequence alone, pointing to the potential for mechanism-oriented multi-modal integration.

---

## 2. Central Dogma Transformer

### 2.1 Design Principles

The Central Dogma Transformer (CDT) is a multi-modal architecture that integrates DNA, RNA, and protein representations through directional cross-attention layers aligned with the biological flow of genetic information. The architecture is designed not only for predictive accuracy but for biological interpretability—each attention layer produces weights that correspond to defined molecular relationships, enabling hypothesis generation about regulatory mechanisms.

The design of CDT embodies a fundamental principle: the structure of the model should reflect the structure of the biology it seeks to understand. The Central Dogma of molecular biology [1] describes the unidirectional transfer of genetic information from DNA to RNA through transcription, and from RNA to protein through translation. This directional flow is not merely a biochemical pathway but the organizing principle of cellular information processing. CDT embeds this directionality into its computational graph through sequential cross-attention layers (Figure 1).

This architectural alignment between computation and biology serves multiple interconnected purposes. First, it provides an inductive bias that constrains the model to learn representations consistent with known biological information flow rather than arbitrary feature combinations. Neural networks are universal function approximators [22] capable of learning any input-output mapping given sufficient capacity; by structuring the architecture to follow the Central Dogma, we guide learning toward biologically plausible representations. Second, this alignment enables interpretability. In conventional multi-modal integration approaches that concatenate embeddings or use bidirectional attention, the learned representations lack clear biological correspondence—high weights between features may reflect statistical correlations without mechanistic meaning. In CDT, each cross-attention layer corresponds to a defined biological process: the DNA-to-RNA layer models transcriptional regulation, while the RNA-to-Protein layer models translational relationships. Consequently, the learned attention weights may carry biological meaning: elevated attention between a DNA position and a gene indicates that the model associates that genomic region with that gene's regulation—a computationally-derived hypothesis that can be evaluated against experimental data.

A third design principle underlying CDT is modularity. Rather than training a monolithic model from raw sequences, CDT operates as an integration framework that accepts pre-trained embeddings from domain-specific foundation models. This modular design offers practical advantages: it leverages the substantial investments in training large-scale language models for each molecular modality, and it allows CDT to benefit from advances in any individual domain without requiring complete retraining. When improved DNA, RNA, or protein language models become available, they can be incorporated by fine-tuning only the lightweight CDT layers (projection, cross-attention, VCE, and task head).

### 2.2 Input Representations

CDT accepts pre-trained embeddings from domain-specific language models for each biological modality: DNA, RNA, and protein. This design philosophy reflects a practical recognition that training large-scale language models requires enormous computational resources and carefully curated datasets that have already been invested by the research community. Rather than duplicating this effort, CDT leverages the knowledge captured by existing foundation models, treating them as feature extractors that encode domain-specific patterns learned from vast corpora of biological sequences.

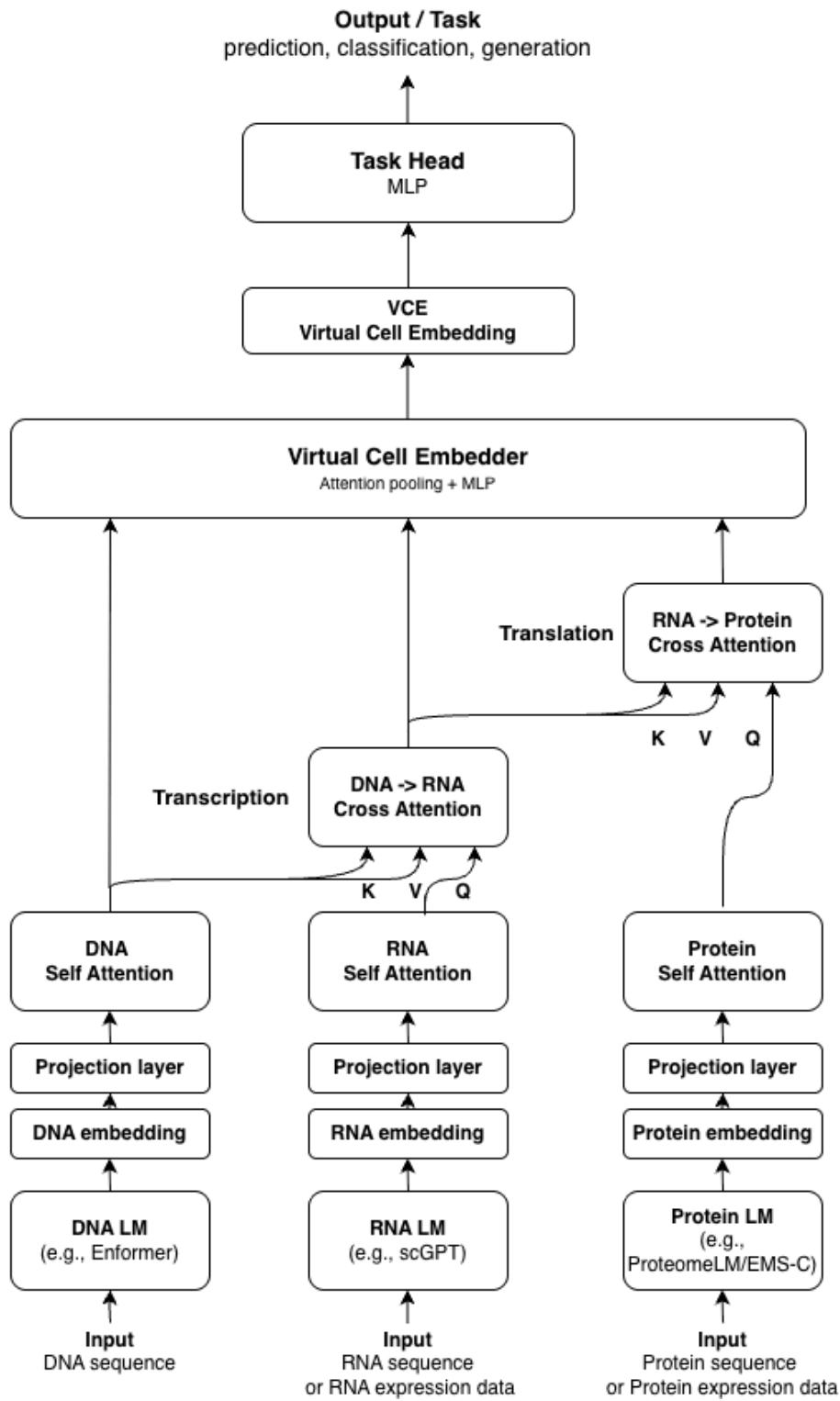

Figure 1: **Central Dogma Transformer Architecture. Input Layer:** Pre-computed embeddings from three foundation models—e.g., Enformer for DNA (genomic context around enhancer regions), scGPT for RNA (gene expression patterns from single-cell data), and ESM-C/ProteomeLM for Protein (sequence embeddings refined with protein-protein interaction context). These can be replaced with other compatible foundation models. **Projection Layer:** Linear projections map each modality's embeddings into a shared 768-dimensional space, enabling cross-modal interactions. **Self-Attention:** Each modality undergoes independent self-attention to capture intra-modal relationships (e.g., long-range genomic interactions, gene co-expression patterns, protein-protein relationships). **Cross-Attention (Central Dogma):** Information flows unidirectionally following the Central Dogma—DNA→RNA cross-attention models transcriptional regulation (which genomic regions influence which genes), and RNA→Protein cross-attention models translational relationships (which transcripts relate to which proteins). This directionality is enforced architecturally, not learned. **Virtual Cell Embedder (VCE):** Attention-based pooling integrates all three modalities into a single cell-state representation, capturing the holistic cellular context. **Task Head:** A prediction layer maps the VCE output to task-specific outputs (e.g., effect size predictions for CRISPRi enhancer perturbations).

The choice of which language model to use for each modality is task-dependent and user-configurable. Different biological questions may benefit from different representations: a task involving long-range genomic regulation might require a DNA language model with extended context windows, while a task focused on coding sequences might prioritize models trained on protein-coding regions. Similarly, RNA representations might come from models trained on bulk RNA-seq, single-cell transcriptomics, or spatial transcriptomics depending on the biological context of interest. This flexibility is a core feature of the CDT framework—the architecture defines how modalities interact, not which specific language models must be used.

Because different language models produce embeddings of different dimensions, CDT employs learned linear projection layers to map each modality's embeddings into a shared internal dimension ($d = 768$ in the current implementation). This design choice simplifies the architecture: with all modalities in the same dimensional space, cross-attention layers and VCE can operate uniformly without handling dimension mismatches. The projection layers are also learnable, allowing the model to adapt pre-trained representations to the specific task during training.

In the current implementation, all foundation models are completely frozen—their parameters remain fixed, and embeddings are pre-computed once and cached to disk. CDT then trains exclusively on these cached representations without requiring the foundation models to be loaded during training. This design is inspired by parameter-efficient fine-tuning approaches such as LoRA [23], which adapt large pre-trained models without modifying their original parameters. By freezing the foundation models and training only the lightweight CDT layers (projection, cross-attention, VCE, and task head), the entire system can be trained quickly with a small number of trainable parameters. The foundation models collectively contain approximately 712 million parameters (Enformer ~250M, scGPT ~50M, ESM-C ~300M, ProteomeLM ~112M), while CDT's trainable layers contain only approximately 60 million parameters—approximately 8% of the foundation model total. This dramatic reduction in trainable parameters enables rapid iteration and experimentation. The approach also addresses practical computational constraints: running Enformer, scGPT, and ProteomeLM simultaneously during training would require substantial GPU memory, while the cached embedding approach enables training on more modest hardware.

For the CRISPRi enhancer prediction task presented in this work, language models were selected based on specific task requirements:

| Modality | Language Model | Dimension | Selection Rationale |
|---|---|---|---|
| DNA | Enformer [3] | 3072 | 114kb window used in this study; pre-trained on chromatin accessibility and histone modification tracks |
| RNA | scGPT [8] | 512 | Gene-level embeddings learned from large-scale single-cell transcriptomic data across diverse cell types |
| Protein | ESM-C/ProteomeLM [11, 12] | 768 | Sequence embeddings refined with protein-protein interaction context |

Table 1: **Foundation Models for CRISPRi Enhancer Prediction Task.**

These selections are not prescriptive for CDT in general. Future applications might substitute Nucleotide Transformer [4] or DNABERT [24] for DNA, Geneformer [9] for RNA, or ESM-2 [11] for protein, depending on the specific biological question and available data. The modular architecture ensures that such substitutions require only retraining of the projection layers and downstream components, not redesign of the integration framework.

## 2.3 Self-Attention Layers

Prior to cross-modal integration, each modality undergoes self-attention processing to capture internal structure within its representation space. Self-attention allows each position within a modality to attend to all other positions, learning pairwise relationships that may reflect biological interactions. This intra-modal processing refines the pre-trained embeddings before they participate in cross-modal attention, allowing the model to emphasize features relevant to the downstream task.

**DNA Self-Attention** (n = 2 layers) operates over the 896 positional embeddings spanning the 114-kilobase genomic window. Each position can attend to every other position, enabling the model to capture long-range dependencies within the DNA sequence context. Biologically, this architecture can represent interactions between distal regulatory elements—for example, an enhancer position attending to a promoter position, or interactions between multiple enhancers that cooperatively regulate gene expression. The attention weights in this layer produce an 896 × 896 matrix that can be examined *post hoc* to identify which genomic positions the model considers related. Two self-attention layers were chosen for DNA to provide sufficient capacity for modeling the complex regulatory grammar of non-coding sequences, where multiple layers of combinatorial logic govern gene expression.

**RNA Self-Attention** (n = 1 layer) operates over gene-level embeddings, where each gene can attend to every other gene in the representation. This architecture can capture co-expression relationships and functional associations between genes. Biologically, genes that participate in the same pathway, respond to the same regulatory signals, or encode proteins in the same complex often show correlated expression patterns. The RNA self-attention layer can learn to represent these relationships, producing an n_genes × n_genes attention matrix that may reflect functional relatedness. A single layer was deemed sufficient given that the gene embeddings from scGPT already encode substantial information about gene relationships learned from large-scale transcriptomic data.

**Protein Self-Attention** (n = 1 layer) operates over protein embeddings, allowing each protein to attend to every other protein. This layer can capture relationships across the proteome, potentially including physical protein-protein interactions, membership in the same functional complex, or participation in the same biological pathway. The attention weights produce an n_proteins × n_proteins matrix that may reflect the modular organization of cellular function. As with RNA, a single layer was used because the input protein embeddings from ProteomeLM already incorporate protein-protein interaction information.

Each self-attention layer employs multi-head attention (h = 8 heads), allowing the model to learn multiple types of relationships simultaneously. Residual connections [25] and layer normalization [26] are applied following standard transformer practice to ensure stable training.

## 2.4 Directional Cross-Attention

The core architectural innovation of CDT consists of two sequential cross-attention layers that implement the Central Dogma information flow [1]. Unlike self-attention, where queries, keys, and values come from the same modality, cross-attention enables one modality to query information from another. The direction of this query—which modality asks and which modality answers—is biologically meaningful and carefully designed to mirror the flow of genetic information.

This directionality distinguishes CDT from conventional multi-modal approaches. In concatenation-based methods, embeddings from different modalities are simply joined into a single vector, erasing the distinction between information sources and making it impossible to trace which modality contributed to a prediction. In bidirectional attention approaches, all modalities attend to all others symmetrically, conflating regulatory relationships with reverse dependencies. CDT's unidirectional cross-attention preserves the asymmetry inherent in biological information flow: DNA informs RNA, and RNA informs protein. This asymmetry is not a limitation but a feature—it ensures that each learned attention weight has a defined biological interpretation.

**DNA → RNA Cross-Attention** models the process of transcriptional regulation. In biological transcription, RNA polymerase reads DNA sequences and synthesizes RNA transcripts; transcription factors bind to regulatory DNA elements (promoters, enhancers, silencers) to modulate which genes are expressed and at what levels. The DNA → RNA cross-attention layer is a computational analog of this process. RNA gene representations serve as queries (Q), while DNA positional representations serve as keys (K) and values (V). This assignment reflects the biological reality that a gene's expression state is determined by which DNA regulatory elements influence it—the gene is the recipient of regulatory information encoded in DNA. Conceptually, each gene "asks" the DNA: "which genomic positions are relevant for my regulation?"—mirroring how the transcriptional machinery interprets DNA regulatory sequences to determine gene expression levels. The attention mechanism computes a relevance score between each gene and each DNA position, producing an n_genes × 896 attention matrix. High attention weights indicate that the model associates a particular genomic region with a particular gene's transcriptional state.

The cross-attention mechanism follows the standard transformer attention formulation:

$$\text{Attention}(Q, K, V) = \text{softmax}\left(\frac{Q \cdot K^T}{\sqrt{d}}\right) \cdot V$$

where $Q$ (query), $K$ (key), and $V$ (value) are learned linear projections of the input embeddings, and $d$ is the embedding dimension for scaling. The softmax normalization ensures that attention weights sum to 1, creating an interpretable distribution over the attended positions.

In DNA → RNA cross-attention, these components are assigned as follows:

- **Q (Query)**: Derived from RNA embeddings — each gene queries for relevant DNA positions
- **K (Key)**: Derived from DNA embeddings — genomic positions provide searchable keys
- **V (Value)**: Derived from DNA embeddings — genomic positions provide information to be aggregated

This produces an n_genes × 896 attention matrix, where each entry represents the relevance of a DNA position for a particular gene. As with self-attention layers, cross-attention employs multi-head attention (h = 8 heads), allowing the model to learn multiple types of regulatory relationships simultaneously —for example, one head might capture promoter associations while another captures distal enhancer relationships.

This layer is particularly powerful for studying gene regulation because the attention weights can be directly examined to identify candidate regulatory elements. If the model has learned biologically meaningful representations, genes should attend strongly to their promoters, enhancers, and other regulatory regions. The attention map thus provides a computational hypothesis about regulatory relationships that can be compared against experimental data from ChIP-seq, ATAC-seq, or chromatin conformation studies.

**RNA → Protein Cross-Attention** models the relationship between transcript and protein levels. In biological translation, ribosomes read mRNA sequences and synthesize polypeptide chains according to the genetic code; the abundance and activity of proteins depend on transcript availability, translation efficiency, and protein stability. The RNA → Protein cross-attention layer captures these dependencies computationally.

In RNA → Protein cross-attention, the components are assigned as follows:

- **Q (Query)**: Derived from Protein embeddings — each protein queries for relevant transcripts
- **K (Key)**: Derived from RNA_fused embeddings — genes provide searchable keys
- **V (Value)**: Derived from RNA_fused embeddings — genes provide information to be aggregated

Here, RNA_fused denotes the output of the preceding DNA → RNA cross-attention layer, not the original RNA input. These fused representations have already integrated information from DNA regulatory regions, carrying forward the genomic context that influences each gene's expression. Thus, when proteins query the RNA layer, they indirectly access DNA-derived regulatory information as well—the information flows sequentially through the Central Dogma pathway, accumulating context at each step.

This assignment reflects the biological reality that a protein's abundance depends on transcript availability and regulation—the protein is the downstream product whose state is influenced by the transcriptome. Each protein "asks" the transcriptome: "which genes' expression levels are relevant for my abundance or function?" The resulting n_proteins × n_genes attention matrix captures associations between proteins and transcripts, learning which transcript features are predictive of each protein's state.

While translation is often viewed as a direct one-to-one mapping from mRNA to protein, the biological relationship between transcript levels and protein abundance is more complex—post-transcriptional regulation, translation efficiency, and protein degradation create many-to-many relationships. The RNA → Protein cross-attention layer can capture these nuanced relationships, learning which transcripts are most predictive of each protein's state. Because this layer receives RNA representations already enriched with DNA-derived regulatory information (via the preceding DNA → RNA layer), it can learn how genomic context influences translation outcomes.

**Unidirectional Information Flow.** CDT enforces strictly unidirectional information flow: DNA → RNA → Protein, with no reverse connections. This design choice prioritizes interpretability over representational flexibility. In biological reality, proteins regulate transcription through feedback loops —transcription factors bind DNA to control gene expression, creating circular causality. However, modeling such feedback computationally would create bidirectional attention patterns that confound interpretation: elevated attention between a DNA region and a gene could reflect either the gene's regulation by that DNA region or the gene product's binding to that region as a transcription factor. By enforcing unidirectional flow, we ensure that each attention weight has a single, defined biological interpretation.

## 2.5 Virtual Cell Embedder

The Virtual Cell Embedder (VCE) is the first component where all three modalities are directly integrated. Prior to VCE, cross-attention layers transfer information sequentially (DNA → RNA → Protein), but each modality remains as a separate representation. VCE receives these cross-informed representations and fuses them into a unified cellular state vector:

- **DNA_emb**: Original DNA positional embeddings (self-attention only, no cross-modal input)
- **RNA_fused**: RNA_emb + DNA information (via DNA → RNA cross-attention)
- **Protein_fused**: Protein_emb + RNA_fused information (via RNA → Protein cross-attention)

Mathematically:

$$\text{RNA}_{\text{fused}} = \text{RNA}_{\text{emb}} + \text{CrossAttention}(Q = \text{RNA}_{\text{emb}}, K = \text{DNA}_{\text{emb}}, V = \text{DNA}_{\text{emb}})$$

$$\text{Protein}_{\text{fused}} = \text{Protein}_{\text{emb}} + \text{CrossAttention}(Q = \text{Protein}_{\text{emb}}, K = \text{RNA}_{\text{fused}}, V = \text{RNA}_{\text{fused}})$$

When VCE merges all three representations, an interesting architectural property emerges: DNA_emb and RNA_fused are input alongside Protein_fused, even though Protein_fused already contains information derived from RNA_fused (which itself contains DNA information). This creates an effect similar to residual connections at the system level—upstream modalities (DNA, RNA) are preserved as direct inputs to VCE, not lost in the sequential flow to Protein. DNA information thus reaches VCE through three paths: directly, through RNA_fused, and through Protein_fused. This architecture ensures that VCE integrates representations at different stages of information accumulation: DNA carries pure genomic information, RNA_fused carries original transcriptomic features plus genomic regulatory context, and Protein_fused carries original proteomic features plus the full Central Dogma pathway context.

The concept of a "Virtual Cell Embedding" reflects a broader vision in computational biology [27]: that the complex molecular state of a cell can be represented as a point in a learned embedding space. Cells with similar functional states should map to nearby points; perturbations that alter cellular function should produce predictable movements in the space. Unlike cell embeddings from single-modality models (e.g., scGPT's cell representations from transcriptomics alone), CDT's VCE integrates information that has flowed through the Central Dogma pathway—each gene's representation carries regulatory context from DNA, and each protein's representation reflects both transcriptomic and genomic influences. This multi-modal, biologically-structured integration distinguishes CDT's VCE from approaches that embed cells based on a single molecular layer.

The VCE employs an attention-based pooling mechanism rather than simple concatenation or mean pooling (see Methods for mathematical formulation). For each modality, a learned query vector (a trainable parameter initialized randomly and optimized during training) attends over the corresponding representations:

- **DNA pooling**: Query attends over 896 positional embeddings → 768-dimensional summary vector
- **RNA pooling**: Query attends over n_genes embeddings → 768-dimensional summary vector
- **Protein pooling**: Query attends over n_proteins embeddings → 768-dimensional summary vector

This selective attention allows the VCE to focus on the subset of genomic positions, genes, and proteins most relevant to the prediction task, rather than treating all features equally. The attention weights from VCE pooling are themselves interpretable, revealing which features the model considers most important for each prediction.

The attended representations from each modality (3 × 768 = 2,304 dimensions) are concatenated and processed through a feed-forward network (two linear layers with GELU activation [28]) to produce the final Virtual Cell Embedding—a fixed-dimensional vector (d = 768) representing the integrated cellular state. This feed-forward fusion enables non-linear combinations of features across modalities, capturing interactions between DNA, RNA, and protein that would not be apparent from any single modality alone.

### 2.6 Task-Specific Prediction Head

The final architectural component is a task-specific prediction head that operates on the Virtual Cell Embedding to produce predictions for the task of interest. The prediction head is intentionally simple—typically a small multi-layer perceptron (MLP)—to ensure that the representational work is performed by the CDT backbone rather than the prediction head. This design encourages the VCE to learn task-relevant representations that capture meaningful biological structure.

For the CRISPRi enhancer-gene regulation task presented in this work, the prediction head consists of two linear layers with GELU activation and dropout regularization (p = 0.3), outputting a scalar value representing the predicted effect size ($\hat{\beta}$) for each enhancer-gene pair. The model is trained with Huber loss (see Methods for mathematical details).

The modular design of CDT separates the task-specific prediction head from the multi-modal integration backbone. This separation enables several practical advantages. First, the same trained CDT backbone could potentially serve multiple prediction tasks by swapping only the prediction head—predicting gene expression levels, perturbation effects, drug responses, or other phenotypes from the same integrated cellular representation. Second, when applying CDT to new tasks, the integration layers may benefit from transfer learning: weights learned from one task could provide useful initialization for related tasks, even if the prediction heads differ. Third, the separation clarifies what each component contributes: the CDT backbone learns to integrate multi-modal biological information, while the prediction head learns the mapping from integrated representation to task-specific output.

## 2.7 Attention Map Interpretation

A distinguishing feature of CDT is that its attention maps may carry biological meaning. In conventional neural networks, internal representations are typically opaque—high-dimensional vectors that capture statistical patterns but resist interpretation. In CDT, the architecture is designed so that attention weights correspond to defined biological relationships: DNA-to-RNA attention represents potential regulatory associations, RNA-to-Protein attention represents potential translational relationships, and within-modality self-attention represents potential interactions among genomic positions, genes, or proteins.

This interpretability arises from the alignment between computational structure and biological structure. When a gene (RNA query) attends strongly to a particular DNA position (key), this indicates that the model considers that genomic region relevant for that gene—a computational hypothesis about gene regulation. When a protein (query) attends strongly to a particular transcript (key), this indicates an association between that protein's state and that gene's expression. These associations are not guaranteed to be causal or mechanistically correct, but they provide starting points for biological investigation.

Table 2 summarizes the attention mechanisms at each layer and their potential biological interpretations:

| Layer | Query | Key/Value | Output Shape | Potential Biological Interpretation |
|---|---|---|---|---|
| DNA Self-Attention | DNA positions | DNA positions | $896 \times 896$ | Interactions between genomic loci; enhancer-promoter relationships; chromatin domain structure |
| RNA Self-Attention | Genes | Genes | $n_{\text{genes}} \times n_{\text{genes}}$ | Gene co-regulation; functional modules; pathway membership |
| Protein Self-Attention | Proteins | Proteins | $n_{\text{proteins}} \times n_{\text{proteins}}$ | Protein complexes; signaling pathways; functional interactions |
| DNA → RNA Cross-Attention | Genes | DNA positions | $n_{\text{genes}} \times 896$ | Regulatory elements for each gene; enhancers, promoters, silencers |
| RNA → Protein Cross-Attention | Proteins | Genes | $n_{\text{proteins}} \times n_{\text{genes}}$ | Transcript-protein associations; translation regulation |
| VCE Attention | Learned queries | All modalities | Variable | Task-relevant features; which positions, genes, and proteins matter for prediction |

Table 2: **Attention Mechanisms and Biological Interpretations.** Each attention layer in CDT produces interpretable attention weights that can be examined for biological meaning.

The attention maps can be examined at multiple granularities. Individual attention patterns reveal which features the model associates for specific samples—for example, which DNA positions are attended for a particular gene in a particular genomic context. Aggregated attention patterns across many samples reveal systematic relationships—for example, whether certain DNA positions are consistently attended across genes, suggesting shared regulatory elements. However, in architectures with fixed context embeddings (as in CDT's current design with fixed scGPT gene token embeddings), cross-attention patterns may be largely uniform across samples, limiting their utility for distinguishing biological conditions.

More broadly, attention weights require careful interpretation. High attention indicates that the model considers a feature during computation, but does not prove that the feature causally drives the prediction. A position may receive high attention because it provides useful context rather than because it is the mechanistic driver. To distinguish "what the model looks at" from "what drives the prediction," complementary analysis methods are needed.

### 2.8 Interpretation Framework

CDT supports two complementary approaches for understanding what the model has learned: attention analysis and gradient analysis. These approaches answer fundamentally different questions and, as we demonstrate in the Results, often highlight different genomic features—suggesting that both are needed for comprehensive interpretation.

**Attention Analysis** examines the forward pass of the model, asking: "What does the model look at during inference?" By extracting attention weights from each layer, we can identify which DNA positions a gene attends to, which genes a protein attends to, and which features the VCE emphasizes. This analysis is straightforward to implement—attention weights are a direct output of transformer computations—and produces interpretable heatmaps that can be overlaid on genomic coordinates or gene networks. Attention analysis is particularly useful for exploratory investigation: scanning across many samples to identify patterns, comparing attention between conditions, and generating hypotheses about regulatory relationships.

**Gradient Analysis** examines the reverse pass, asking: "What drives the model's predictions?" By computing the gradient of the prediction with respect to input DNA embeddings (see Methods for mathematical formulation), we identify which positions would most change the output if perturbed [19]. This approach is analogous to a computational perturbation experiment: rather than physically altering each genomic position and measuring the effect, we compute the mathematical sensitivity of the prediction to each position.

A key property of gradient analysis in CDT is what we term "reverse Central Dogma tracing": by the chain rule, gradients flow backward through the architecture—from prediction through VCE, through RNA-to-Protein cross-attention, through DNA-to-RNA cross-attention, and ultimately to specific DNA positions. This computational pathway mirrors the reverse of biological information flow, enabling attribution from phenotype back to genomic sequence.

Empirically, attention and gradient analyses often highlight largely non-overlapping genomic regions. This divergence is not a failure of either method but reflects their different questions: attention reveals the broad context the model integrates, while gradients identify the specific features critical for each prediction. A genomic position may receive high attention because it provides useful contextual information, even if perturbing that position would not change the prediction. Conversely, a position may have high gradient despite low attention if it exerts strong influence through indirect pathways. The complementarity of these methods suggests a practical workflow: attention analysis for broad exploration and hypothesis generation, gradient analysis for identifying causally relevant features for experimental follow-up. A detailed comparison of these approaches, with case studies demonstrating their complementary utility, is presented in the Results and Discussion sections.

# 3. Results

## 3.1 Gene Selection: Addressing Cross-Experiment Variability

A key challenge in integrating multi-modal data is that different experiments capture different aspects of cellular state. The Gasperini et al. CRISPRi screen [29] and the Morris et al. STING-seq experiment [30] both profiled K562 cells, but their scRNA-seq data showed substantial differences—reflecting technical variability inherent to single-cell transcriptomics.

To quantify this variability, we compared mean gene expression levels between the two K562 experiments across 10,492 common genes. Despite profiling the same cell line, the Pearson correlation was $r = 0.797$ (Spearman $r = 0.91$), reflecting substantial batch effects inherent to independent scRNA-seq experiments. This baseline variability sets an upper bound on achievable prediction accuracy: no model trained on one experiment's data can perfectly predict another experiment's measurements, because the ground truth itself varies between experiments.

To ensure robust analysis, we selected genes meeting three criteria: (1) detected as expressed in both K562 experiments, (2) present in scGPT vocabulary, and (3) present in ESM-C/ProteomeLM. This intersection yielded 2,360 genes that are robustly expressed across independent experiments and have both RNA and protein representations available.

## 3.2 CDT Predicts Enhancer Effects from Sequence

To evaluate whether multi-modal integration can predict enhancer effects, we trained CDT on CRISPRi perturbation data from the Gasperini et al. dataset [29]. The dataset contains 4,605 enhancer-gene pairs for training and 996 pairs for validation, split at the enhancer level by stratified random sampling based on effect significance (positive/negative labels). This splitting strategy ensures that each enhancer appears exclusively in either training or validation, preventing information leakage from DNA sequences seen during training. Each pair is annotated with a beta value representing the effect size coefficient from the original analysis (natural log scale; negative values indicate that enhancer silencing reduces gene expression). We pre-computed embeddings from frozen foundation models for DNA (Enformer), RNA (scGPT), and protein (ESM-C/ProteomeLM), with all language models frozen during CDT training (see Methods for detailed embedding preparation and cross-modal alignment procedures). Notably, unlike typical Enformer applications that center sequences on transcription start sites (TSS), we centered DNA sequences on enhancer positions to capture the genomic context surrounding each regulatory element.

The experimental design ensures that enhancer DNA sequence is the only input that varies across samples; RNA and protein embeddings are shared across all samples, providing gene and protein representations as contextual information. The model receives a target gene index to specify which gene's effect to predict, and this index is used to select the corresponding RNA and protein embeddings for that gene. However, these embeddings represent pre-trained gene and protein representations, not sample-specific information about effect sizes—the model must learn to predict enhancer effects from the varying DNA sequence features. CDT achieved a Pearson correlation coefficient ($r$) of 0.503 on the held-out validation set (Figure 2A). Given the cross-experiment variability ($r = 0.797$), this represents approximately 63% of the theoretical ceiling, demonstrating effective extraction of biological signal from inherently noisy experimental data. Training converged stably over 26 epochs (Figure 2B), with validation performance reaching a plateau (training $r = 0.65$, validation $r = 0.503$). The gap between training and validation performance indicates moderate overfitting. This correlation ($r = 0.503$, $R^2 = 0.25$) indicates that the model explains approximately 25% of the variance in enhancer effects—a moderate but meaningful level of prediction from sequence information. Having established that CDT captures predictive signal, we next examined what patterns the model learned and whether the learned representations yield biological insight.

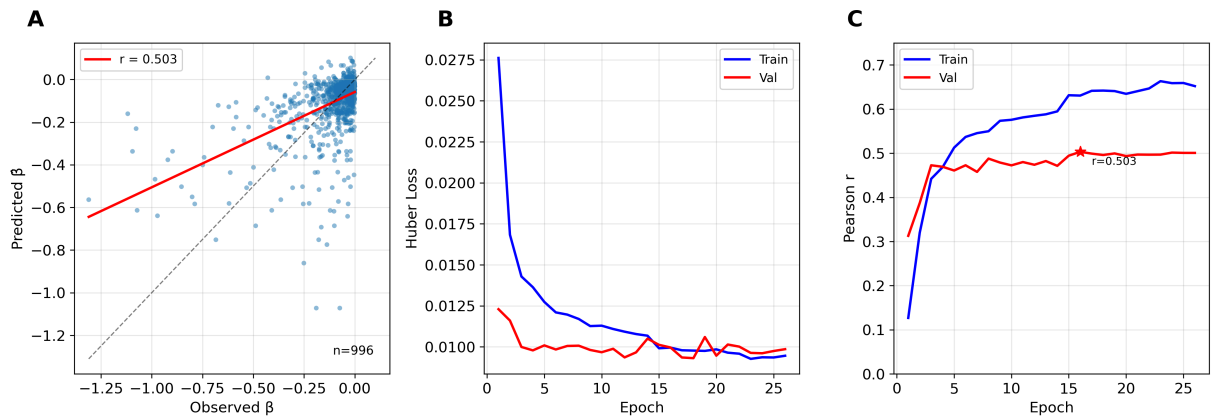

Figure 2: **CDT Prediction Performance. (A)** Scatter plot of predicted versus observed enhancer effect sizes (β) on the held-out validation set (n = 996). Red line shows the linear regression fit (Pearson r = 0.503, p < $10^{-64}$). **(B)** Training loss (Huber loss) over epochs. Blue: training set; Red: validation set. **(C)** Pearson correlation over epochs. The model achieved best validation performance (r = 0.503) at epoch 16, with early stopping at epoch 26. The gap between training and validation curves indicates moderate overfitting.

## 3.3 Cross-Attention Reveals Spatial Patterns in DNA-Gene Relationships

We first examined the DNA-to-RNA cross-attention maps, which indicate which genomic positions the model associates with each gene's regulation. In CDT's architecture, this cross-attention layer allows RNA representations (as queries) to attend to DNA positional embeddings (as keys and values), learning which genomic regions are relevant for each gene's transcriptional state. Because this attention mechanism is computationally analogous to the biological process of transcriptional regulation—where DNA regulatory elements influence gene expression—the learned attention weights may be interpretable as computational representations of regulatory associations, though this interpretation requires validation. Analysis across 100 randomly selected validation samples revealed consistent spatial patterns: attention peaks showed a mean distance of approximately 30 kilobases from enhancer centers, with 82% of peaks falling within 50 kilobases (Figure 3). This distance range is consistent with typical enhancer-promoter separations in the human genome, where most regulatory interactions occur within topologically associating domains (TADs) [31]. The model learned these distance relationships without explicit supervision, suggesting that the pre-trained Enformer embeddings encode positional information that CDT leverages to associate regulatory elements with their target genes.

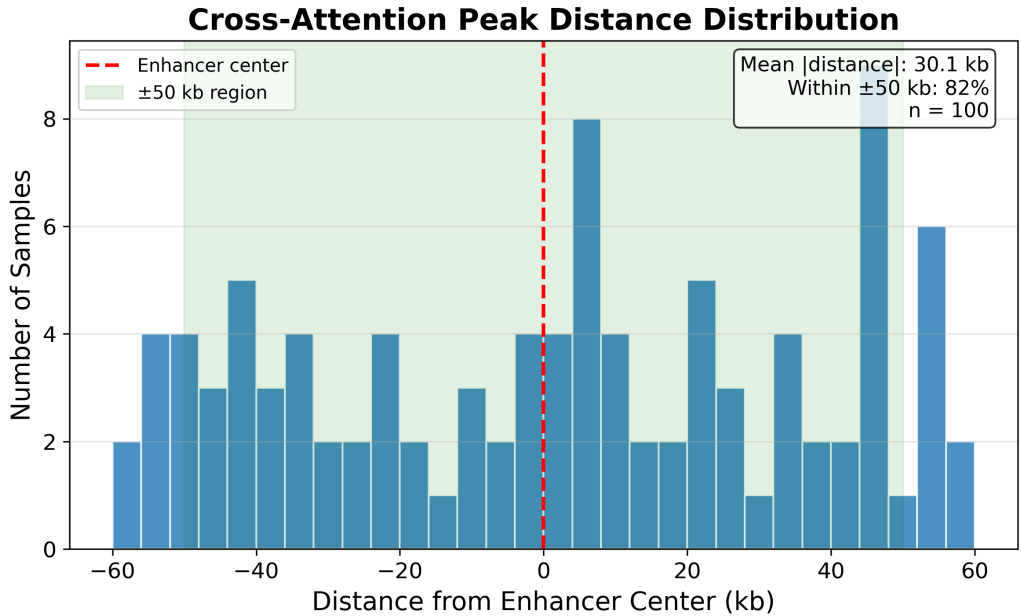

Figure 3: **Cross-Attention Peak Distance Distribution.** Histogram showing the distance of attention peaks from enhancer centers across 100 randomly selected validation samples. Red dashed line indicates the enhancer center. Green shaded region shows the ±50 kb range. Mean absolute distance: 30 kb; 82% of peaks fall within ±50 kb of the enhancer center, consistent with typical enhancer-promoter distances within topologically associating domains (TADs).

Importantly, cross-attention patterns were highly uniform across samples. Because RNA embeddings are pre-trained scGPT gene token embeddings (identical across all samples), the Query vectors in DNA→RNA cross-attention do not vary between samples. The model "looks at" similar DNA positions regardless of sample-specific effect sizes. This architectural property limits the utility of cross-attention for distinguishing individual regulatory mechanisms, but does not diminish the model's predictive power —the sample-specific information flows through DNA embeddings and is captured by downstream processing. This design reflects our proof-of-concept approach using CRISPRi data, where we prioritized avoiding data leakage over capturing gene-specific expression patterns. Future extensions incorporating per-cell RNA embeddings from single-cell experiments could enable more detailed cross-attention patterns that reflect cell-state-specific regulatory relationships. To identify what drives individual predictions in the current architecture, we turned to gradient analysis.

### 3.4 Gradient Analysis Reveals Prediction-Critical Features Distinct from Attention

To understand what drives CDT's predictions, we employed gradient analysis—a technique that identifies which input features most strongly influence the model's output. While attention analysis reveals what the model "looks at" during inference, gradient analysis answers a different question: "If we changed this input feature slightly, how much would the prediction change?" The gradient of a neural network output with respect to an input position measures the sensitivity of the prediction to that position; a large gradient magnitude indicates that small changes at that position would substantially alter the prediction. In biological terms, gradient analysis performs a computational perturbation experiment: it identifies genomic positions where sequence variation would most affect the predicted enhancer effect, without requiring actual experimental perturbation. This provides a form of counterfactual reasoning —highlighting positions that are computationally linked to the prediction through the model's learned function, though whether this computational linkage reflects true biological mechanism requires external validation.

The biological interpretability of gradient analysis depends on the model architecture. In a blackbox neural network, high gradients indicate computational sensitivity but carry no inherent biological meaning. In CDT, however, the architecture is designed to mirror the biological Central Dogma: information flows from DNA through RNA to protein, and when we compute gradients (which flow in the reverse direction during backpropagation), they trace back through this pathway. A high gradient at a DNA position indicates that this genomic region strongly influences the predicted phenotype through the model's computation—whether this reflects genuine biological regulatory mechanisms remains to be established through experimental validation.

We found that attention and gradient analyses identified largely non-overlapping genomic regions. When we compared the top-20 positions ranked by attention weights versus those ranked by gradient magnitude across 100 validation samples, the mean overlap was only 2.2 out of 20 DNA positions (approximately 10%) (Figure 4). This divergence suggests a distinction between what the model considers during computation (attention) and what computationally drives predictions (gradients)—a position can have high attention but low gradient, or vice versa. The gradient magnitude distribution also revealed an asymmetry across modalities: DNA positions showed larger gradient magnitudes than RNA and protein features, reflecting the task design where DNA embeddings vary across samples while RNA and protein embeddings are fixed pre-trained representations. This divergence between attention and gradient raised a question: if they highlight different regions, which better reveals biological mechanism? To address this, we examined the FNDC5 sample in detail—selected because it exhibited the largest effect size ($\beta$) among validation samples—where attention and gradient pointed to different genomic positions.

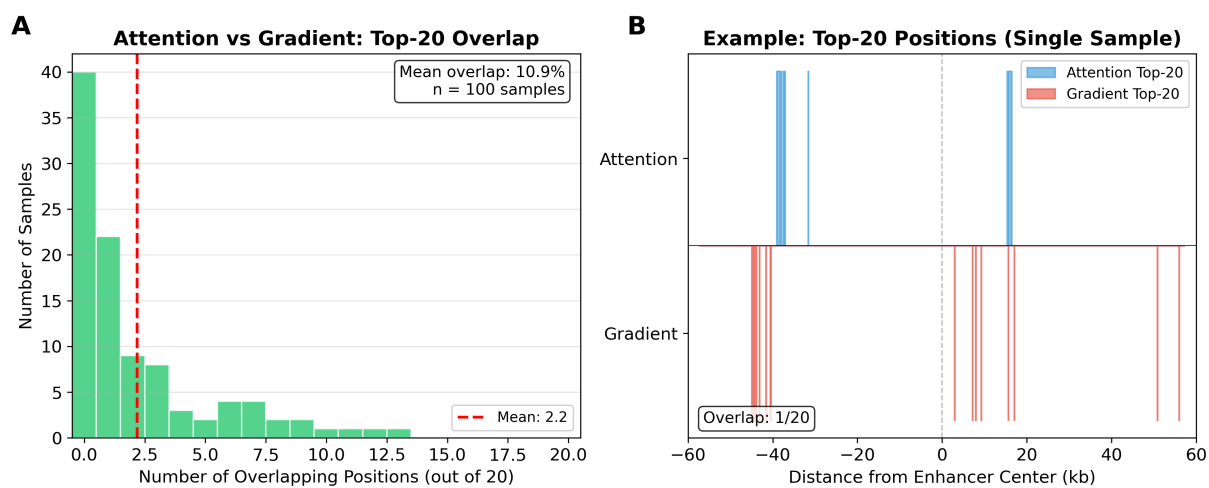

Figure 4: **Attention and Gradient Identify Distinct Genomic Regions. (A)** Distribution of overlap between top-20 attention and top-20 gradient positions across 100 validation samples. Mean overlap is 2.2 positions (approximately 10%), indicating that these methods identify largely non-overlapping regions. **(B)** Example from a single sample showing the spatial distribution of top-20 positions identified by attention (blue, top) versus gradient (red, bottom). Positions are shown relative to enhancer center.

### 3.5 FNDC5 Case Study: Following Gradients to a Functional Element

To illustrate the application of gradient analysis for mechanistic discovery, we selected the sample with the strongest enhancer effect in the validation set: sample 3449 targeting FNDC5 (beta = −1.31, ranked 1st among 996 validation samples). This case exemplifies the scientific workflow enabled by CDT: computational analysis generates a specific, testable hypothesis about regulatory mechanism, which can then be validated against independent experimental data. The enhancer in this sample is located at chr1:33,592,642-33,593,311, positioned within an intron of the CSMD2 gene. The target gene is FNDC5, which encodes irisin—a myokine released during exercise [32] that has attracted considerable research interest for its role in metabolism and potential therapeutic applications. While FNDC5 transcriptional regulation has been studied at the promoter level—including PGC-1α/ERRα-mediated exercise response and glucocorticoid receptor regulation [32]—the role of distal enhancers and CTCF-mediated chromatin architecture in FNDC5 regulation has not been previously characterized. FNDC5 lies at chr1:32,862,268-32,870,912, approximately 726 kilobases away from the enhancer (center-to-center distance). This distance poses a fundamental question: how can an enhancer regulate a gene nearly three-quarters of a megabase away? The CRISPRi experiment showed strong regulatory effect (beta = −1.31, indicating substantial gene expression reduction upon enhancer perturbation), confirming that this distal enhancer genuinely regulates FNDC5. But the mechanism of this long-range regulation remained unclear.

We first examined the cross-attention map for this sample to understand which genomic positions the model associated with FNDC5 regulation. The attention analysis identified a peak at +47.4 kilobases relative to the enhancer center (Figure 5A). Querying this position against ENCODE annotations revealed candidate regulatory elements: a CTCF binding site (E1336104) and a distal enhancer-like element (E1336105/enhD). The functional connection of these elements to FNDC5 regulation remains to be experimentally validated. However, since this position lies only 47kb from the enhancer, it likely resides within the same topologically associating domain (TAD) as the enhancer and FNDC5—suggesting potential spatial proximity in three-dimensional nuclear space that warrants future investigation. We then performed gradient analysis, computing the sensitivity of the predicted FNDC5 effect to each input position. The gradient analysis identified a different peak: −56.7 kilobases from the enhancer center, at genomic coordinate chr1:33,536,296 (Figure 5B). This position showed the highest gradient magnitude among all 896 DNA positions in the input window—indicating that, according to the model's learned function, this is the genomic position whose modification would most strongly affect the predicted enhancer effect on FNDC5.

We queried this gradient-identified position against the ENCODE database of regulatory elements and found that chr1:33,536,296 falls within a CTCF binding site region, catalogued as ENCODE element E1308103 (note: the 128-bp bin resolution of our analysis means the exact nucleotide position is approximate). This was not information available to the model during training—CDT was trained only to predict enhancer effects from sequence embeddings, with no explicit supervision about CTCF or chromatin architecture. CTCF (CCCTC-binding factor) [20] is a major regulator of three-dimensional genome organization, functioning as an architectural protein that defines the boundaries of chromatin loops and topologically associating domains (TADs) [31]. When CTCF binds to specific DNA sequences, it can anchor chromatin loops that bring distant genomic regions into physical proximity, enabling regulatory interactions across large linear distances. The canonical model of long-range enhancer regulation involves CTCF-mediated loops that position enhancers near their target gene promoters in three-dimensional nuclear space, despite being separated by hundreds of kilobases in the linear genome. The gradient analysis thus generated a specific mechanistic hypothesis about this long-range regulation (Figure 5C). Notably, the CTCF site (-56.7kb from enhancer) lies in the same direction as FNDC5 (-726kb from enhancer). Rather than serving as a direct bridge between enhancer and gene, this CTCF site likely functions as a TAD boundary or loop anchor that defines a chromatin domain containing both the enhancer and the distant FNDC5 promoter. By constraining chromatin interactions within this domain, CTCF-mediated architecture would bring the enhancer into spatial proximity with FNDC5 despite their 726kb linear separation. This hypothesis is testable: if correct, Hi-C data should show physical contact between the enhancer region and the FNDC5 locus.

Several aspects of this finding merit emphasis. First, the discovery was blind: we selected this sample solely because it had the strongest enhancer effect in the validation set, and only discovered through gradient analysis that this position corresponds to a CTCF binding site annotated in ENCODE. Second, the model received no explicit training on CTCF binding or chromatin architecture—it was trained solely to predict CRISPRi effect sizes from sequence embeddings, suggesting that Enformer's pre-trained embeddings encode features predictive of CTCF binding and that CDT learned to leverage these features. Third, this demonstrates why both attention and gradient analysis are needed: Cross-attention peaked at +47.4kb, where ENCODE annotations exist (CTCF, enhD), but these elements lack validated connection to FNDC5. Notably, the +47.4kb position—only 47kb from the enhancer—likely resides within the same TAD as the enhancer and FNDC5 (see Hi-C analysis below), raising the possibility that this attention-identified region may also be spatially proximal to FNDC5 in three-dimensional nuclear space. The functionally relevant CTCF site emerged only through gradient analysis, which identified −56.7kb as the position most critical for the prediction—a site subsequently validated by Hi-C data showing contact with FNDC5.

While Enformer's pre-training on CTCF ChIP-seq data provides the capability to recognize CTCF binding motifs, CDT's contribution through gradient analysis was identifying which specific CTCF site among many in the 114kb window is functionally relevant for this particular enhancer-gene relationship. This case study transforms an abstract prediction (beta = −1.31) into a concrete mechanistic hypothesis about three-dimensional chromatin organization, which the next section tests against independent experimental data.

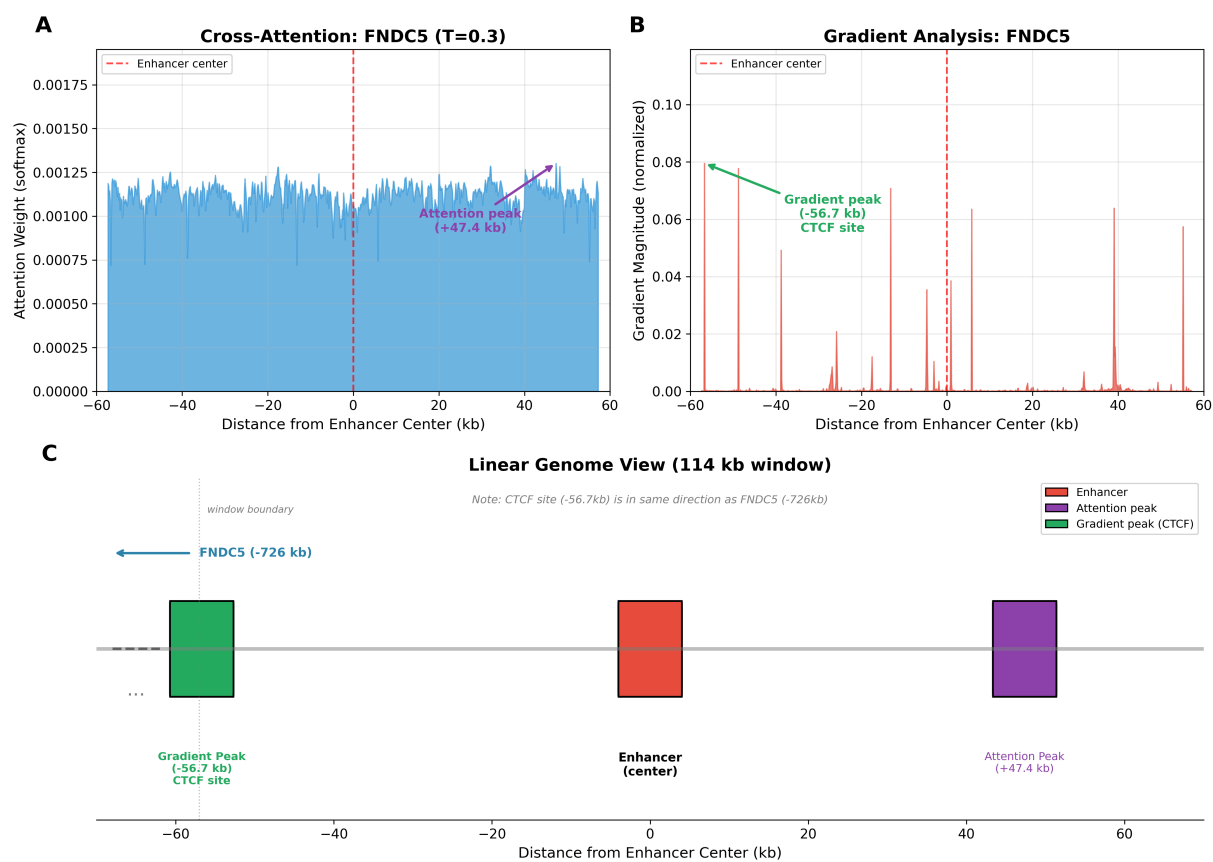

Figure 5: **FNDC5 Case Study: Attention and Gradient Analysis Reveal Distinct Regulatory Elements.** **(A)** Cross-Attention map showing the strongest enhancer effect sample (FNDC5, beta = −1.31). Temperature scaling (T=0.3) was applied to sharpen the attention distribution for visualization; raw attention weights are nearly uniform across positions, but scaling reveals relative differences. Attention peaks at +47.4kb from enhancer center. **(B)** Gradient analysis of the same sample reveals a different critical position at −56.7kb, corresponding to an annotated CTCF binding site. **(C)** Linear genome view showing the spatial relationship between the attention peak (+47.4kb), gradient peak/CTCF site (-56.7kb), enhancer, and FNDC5 gene location (-726kb). Note that the CTCF site lies in the same direction as FNDC5, consistent with a chromatin looping mechanism.

## 3.6 Hi-C Data Shows Three-Dimensional Chromatin Contact

The FNDC5 case study generated a specific, testable prediction: if the CTCF site identified by gradient analysis truly mediates the enhancer-gene interaction, then the enhancer and FNDC5 should be in physical proximity in the three-dimensional nuclear space, despite their 726-kilobase linear separation. We tested this prediction using publicly available Hi-C data from K562 cells (accessed via the 3D Genome Browser). Hi-C [21] is a chromosome conformation capture technique that maps physical contacts between genomic regions across the entire genome by crosslinking DNA regions that are spatially proximate in the nucleus, regardless of their distance along the linear chromosome. The resulting contact frequency matrix reveals the three-dimensional organization of chromatin, providing an independent experimental readout completely separate from the sequence-based predictions of CDT. For our mechanistic hypothesis to hold, two conditions should be met: (1) the enhancer region and FNDC5 should show elevated contact frequency despite their 726kb separation; and (2) both regions should occupy compatible chromatin compartments that permit regulatory interaction.

Analysis of K562 Hi-C data revealed elevated contact signal between the enhancer region (chr1:33.59Mb) and the FNDC5 promoter region (chr1:32.87Mb) (Figure 6). This contact frequency is notable: most genomic regions separated by 726 kilobases show minimal interaction, as contact probability decays rapidly with linear distance. The observed contact indicates that these regions are brought into spatial proximity through active chromatin looping mechanisms—precisely the mechanism our gradient analysis implicated. Examination of the broader chromatin architecture revealed that both the enhancer and FNDC5 reside within the same Topologically Associating Domain (TAD) [31], fundamental units of chromatin organization within which genomic regions interact more frequently with each other than with regions outside the domain. TAD boundaries are often defined by CTCF binding sites [20]—connecting directly to our gradient analysis finding. We further examined the chromatin compartment status using the first principal component (PC1) of the Hi-C contact matrix, finding that both the enhancer region and FNDC5 showed positive PC1 values, indicating they reside in the transcriptionally active A compartment, as expected for functional enhancer-gene pairs. Together, these findings support a coherent mechanistic model (Figure 6): the enhancer contains regulatory sequences that influence FNDC5 expression; a CTCF site at −56.7kb from the enhancer serves as a chromatin loop anchor; CTCF-mediated looping brings the enhancer into spatial proximity with the FNDC5 promoter; and this three-dimensional proximity enables regulation despite 726kb linear separation, with both regions residing in the active A compartment within a shared TAD context.

A notable aspect of this validation is what CDT did not have access to during training. The model was trained solely on DNA sequence embeddings (from Enformer), RNA expression embeddings (from scGPT), and protein embeddings (from ProteomeLM). It received no Hi-C data, no CTCF ChIP-seq data, no explicit information about three-dimensional chromatin organization. Yet gradient analysis identified a CTCF site as the most prediction-relevant genomic position, and Hi-C data confirmed that the enhancer and FNDC5 are in physical contact within the same TAD—consistent with CTCF-mediated chromatin looping. This suggests that Enformer's pre-trained embeddings may encode features predictive of CTCF binding and chromatin architecture, and that CDT learned to leverage these features for enhancer effect prediction. The Central Dogma-aligned architecture may have facilitated this learning: by structuring information flow to parallel biological information flow, CDT was able to capture regulatory relationships that depend on three-dimensional genome organization, even without explicit training on chromatin conformation data. We note important caveats: this is a single case study, and systematic validation across many enhancer-gene pairs would strengthen the conclusion; the Hi-C data shows spatial proximity but does not prove that the CTCF site causally mediates the interaction—perturbation of the CTCF site would provide more direct evidence; and the gradient analysis identifies the most prediction-sensitive position, but this sensitivity could arise from multiple features encoded at that position, not solely CTCF binding.

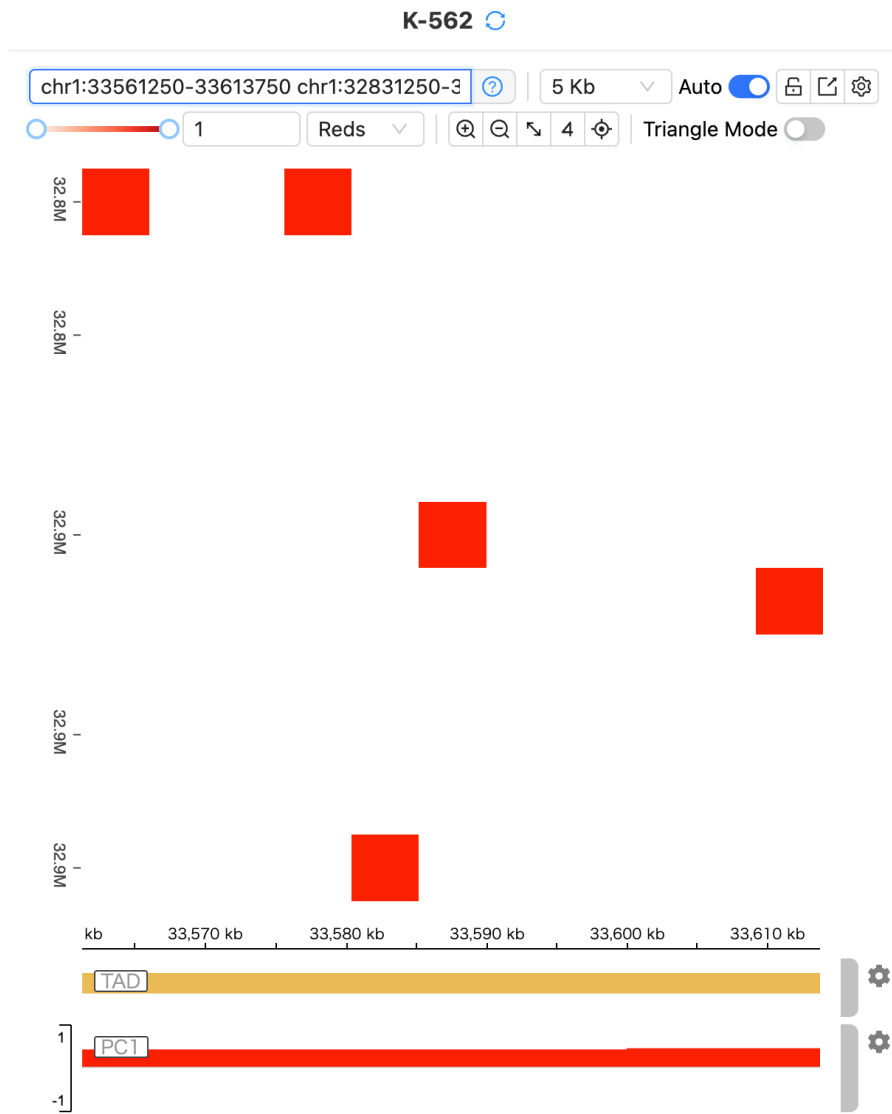

Figure 6: **Hi-C Validation of Enhancer-FNDC5 Chromatin Contact.** K562 Hi-C contact map from 3D Genome Browser showing the interaction between the enhancer (chr1:33.59Mb) and FNDC5 promoter (chr1:32.87Mb). Red squares indicate elevated contact frequency between these regions despite 726kb linear separation. Bottom tracks show: TAD annotation confirming both regions are within the same topologically associating domain; PC1 values are positive (red), indicating both regions reside in the transcriptionally active A compartment. Data source: K562 in situ Hi-C (PRJNA380394).

Despite these caveats, the convergence of gradient analysis and Hi-C validation provides evidence that CDT may have learned biologically meaningful representations of chromatin architecture.

### 3.7 Two Complementary Windows into Regulatory Mechanism

The analyses presented above reveal that CDT provides two distinct and complementary windows into regulatory mechanism. Attention analysis reveals what the model considers during inference—spatial patterns and the features the model incorporates into its representations. Gradient analysis reveals what drives predictions—the specific input positions whose modification would most change the output, tracing back through the Central Dogma-aligned computational graph. The minimal overlap between attention and gradient top positions observed in this case is not a limitation but rather an insight: these approaches may capture fundamentally different aspects of model behavior. Attention represents the broad context the model integrates, while gradients identify the specific features critical for each prediction. In the FNDC5 case, attention highlighted a position (+47.4kb) whose functional connection to FNDC5 remains unvalidated, while gradient analysis pinpointed a CTCF binding site (-56.7kb) subsequently validated by Hi-C data. However, an intriguing hypothesis emerges: since Hi-C confirmed that the enhancer and FNDC5 reside within the same TAD and show three-dimensional proximity, the attention peak at +47.4kb—only 47kb from the enhancer—likely also falls within this TAD. If so, this position may also be spatially proximal to FNDC5 in three-dimensional nuclear space, suggesting that the attention-identified region could have biological relevance that warrants experimental investigation.

This reframing transforms the apparent divergence between attention and gradient into a strength: rather than one method being "right" and the other "wrong," both may illuminate different aspects of three-dimensional chromatin organization within the same TAD. Attention identifies regions the model broadly associates with the regulatory relationship (+47.4kb), while gradient pinpoints the specific feature most critical for prediction (-56.7kb). Together, they may reveal multiple regulatory elements within a chromatin domain that collectively contribute to gene regulation—providing complementary windows into cellular architecture that neither approach could offer alone. This complementarity suggests a practical workflow: attention analysis for broad exploration of what the model has learned, and gradient analysis for identifying causally relevant features for experimental follow-up.

Because CDT's architecture mirrors the biological Central Dogma, both interpretation approaches carry mechanistic meaning. Attention maps in the DNA-to-RNA layer correspond to transcriptional regulatory relationships; those in the RNA-to-Protein layer correspond to translational relationships. Gradients trace prediction-relevant information backward through these same biological layers, identifying which DNA positions influence the predicted phenotype through the transcription-translation cascade. This tight coupling between computational structure and biological structure enables CDT to generate mechanistic hypotheses—predictions about which genomic features regulate which genes through which molecular pathways—that are directly testable through experimental validation.

### 3.8 Attention Head Diversity: Challenges in Interpretation

While attention maps provide interpretable windows into model behavior, multi-head attention introduces interpretive complexity. CDT uses 8 attention heads in each layer, and different heads learn markedly different patterns. Figure 7 illustrates this diversity in DNA Self-Attention: some heads show concentrated attention on specific positions, while other heads show diffuse attention distributed across the full 114kb window. This head diversity is not a bug but a feature of multi-head attention—different heads specialize in capturing different types of relationships. However, it raises a fundamental interpretive question: which head should we examine when seeking biological insights?

This question connects to an ongoing debate in the broader machine learning community regarding whether attention weights can serve as explanations. The seminal "Attention is not Explanation" [33] argued that attention distributions do not reliably indicate feature importance, while "Attention is not not Explanation" [34] countered that interpretability depends on model architecture and task context. In CDT's case, the diffuse attention patterns observed across many heads suggest that determining which positions are biologically meaningful requires additional analysis techniques beyond examining raw attention weights.

Several approaches have been proposed to make attention more interpretable: attention regularization to enforce sparsity [35], semi-supervised attention using domain heuristics, and fully supervised attention trained with human annotations or known biological signals. Guided attention mechanisms [36] can incorporate external supervision directly on attention maps during training. Applying such techniques to CDT—for example, using known enhancer-gene interactions to supervise cross-attention—represents a promising direction for future work.

In the current architecture, gradient analysis provides a complementary, head-invariant perspective. Rather than asking what the model attends to (which varies by head), gradients ask what drives the prediction (which aggregates across all heads). The FNDC5 case study exemplifies this complementarity: attention analysis showed patterns that varied substantially by head, but gradient analysis consistently identified the CTCF site at −56.7kb as prediction-critical across all analytical approaches.

DNA→RNA cross-attention presents a different challenge: patterns are remarkably uniform across genes. Analysis revealed near-identical attention profiles for all genes (pairwise correlation > 0.99), regardless of their biological function or expression level. This uniformity arises from two architectural constraints: (1) the RNA embeddings are pre-trained scGPT gene token embeddings, providing identical query vectors for all samples, and (2) the DNA input covers only the enhancer-centered 114kb window, without direct representation of gene TSS regions. Consequently, cross-attention is dominated by DNA positional features from the Enformer embeddings rather than learning gene-specific regulatory relationships. The cross-attention visualization in Figure 5A, which required temperature scaling (T=0.3) to reveal any positional preferences, illustrates this uniformity. Future extensions incorporating dual DNA windows (enhancer-centered and gene TSS-centered) may enable explicit learning of enhancer-gene spatial relationships.

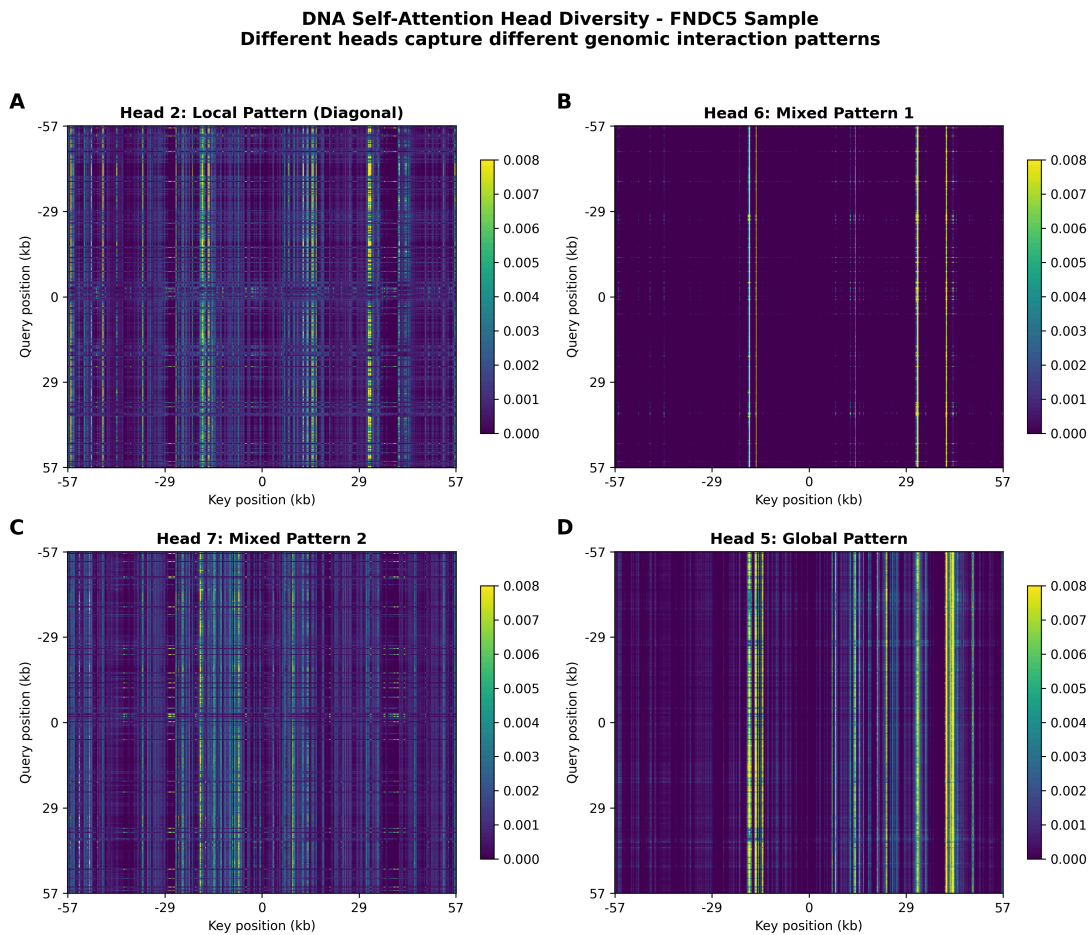

Figure 7: **Attention Head Diversity in DNA Self-Attention.** Four representative heads from Layer 1 showing distinct attention patterns. (A) Concentrated attention on specific positions. (B-C) Mixed patterns with varying focus. (D) Diffuse attention distributed across the 114kb window. This diversity makes interpretation challenging: different heads capture different aspects of genomic relationships, and averaging may obscure biologically meaningful patterns.

# 4. Discussion

## 4.1 FNDC5 Case Study: Potential Chromatin Architecture Signal

While CDT achieved moderate prediction accuracy (Pearson r = 0.503), detailed examination of a single case study suggests the model may have captured features related to chromatin organization. We emphasize that this finding derives from one sample (FNDC5) and requires validation across additional cases before drawing general conclusions.

In the FNDC5 case, gradient analysis identified position −56.7kb as the most prediction-critical among 896 DNA positions. This position overlaps with an annotated CTCF binding site (ENCODE E1308103). Hi-C data independently showed physical contact between the enhancer region and the target gene 726 kilobases away, with the CTCF site positioned to potentially anchor this interaction.

This finding is notable because CTCF binding motifs are abundant—approximately 55,000 sites genome-wide, with likely multiple motifs within any 114kb window. Yet gradient analysis identified this specific position as the most prediction-critical among all 896 DNA positions.

Two factors may contribute to this result. First, Enformer was trained on CTCF ChIP-seq data among thousands of genomic tracks, so its embeddings likely already encode CTCF binding site information. Second, CDT was trained on CRISPRi perturbation effects, which may have enabled learning which genomic positions are relevant for predicting enhancer effects. However, we have not separated these contributions—whether the CTCF site's prediction-critical status derives primarily from Enformer's pre-training, CDT's task-specific learning, or their combination remains unclear.

We note that this finding requires experimental validation to establish causality—Hi-C shows spatial proximity and gradient analysis identifies the CTCF site as prediction-critical, but neither proves causal mediation. Perturbation experiments (CRISPR deletion of the CTCF site) would be needed to test this hypothesis (see Limitations).

The FNDC5 case study represents one detailed examination; whether similar patterns exist in other samples remains unknown. Systematic gradient analysis across all validation samples would be needed to assess the generality of chromatin architecture signals in CDT's predictions.

## 4.2 Two Windows: A Methodological Framework

A key methodological contribution of this work is the distinction between attention analysis and gradient analysis as complementary but fundamentally different approaches to model interpretation. Across 100 validation samples, these approaches identified largely non-overlapping genomic regions (approximately 10% overlap for DNA positions), suggesting they answer different questions about model behavior.

Attention analysis reveals what the model considers during inference—which positions contribute information to the computation. However, the relationship between attention weights and prediction importance is not straightforward. A position may receive high attention because it provides useful contextual information for computing representations, even if that position has minimal influence on the final prediction. This limitation has been recognized in the broader machine learning literature, where studies have shown that attention weights do not reliably indicate which inputs causally drive model outputs [33]. In biological terms, a genomic region might be "consulted" by the model for context without being the regulatory element that determines the predicted effect.

Gradient analysis addresses a different question: which input features, if modified, would most change the prediction? This is a counterfactual question—gradients measure sensitivity to hypothetical pertur-bations. In CDT, gradient computation has an additional structure that distinguishes it from gradient analysis in arbitrary neural networks. Because CDT's architecture mirrors the Central Dogma, gradients flow backward through layers that correspond to biological information transfer: from prediction, through protein-level computation, through RNA-level computation, to DNA positions. We term this "reverse Central Dogma tracing." The gradient at a DNA position reflects how strongly that position influences the prediction through its propagated effects on transcription and translation—the same causal chain that operates in biological gene regulation. This architectural alignment means that gradient peaks in CDT are not arbitrary computational sensitivities; they identify positions whose influence propagates through biologically-structured pathways. Whether this structure genuinely captures biological causation, however, requires experimental validation rather than architectural assumption.

The FNDC5 case illustrated the potential value of this distinction. Attention analysis identified a peak at +47.4kb within a CSMD2 intron—a position with ENCODE-annotated regulatory elements (CTCF, enhD). Since this position lies only 47kb from the enhancer, it likely resides within the same TAD as the enhancer and FNDC5, suggesting potential biological relevance through three-dimensional proximity that warrants future experimental investigation. Gradient analysis identified a different position (-56.7kb) as prediction-critical, and this position corresponded to a CTCF binding site that Hi-C data associated with chromatin contact between the enhancer and FNDC5. While the model "looked at" one region and was "driven by" another, both positions may be biologically meaningful through TAD-mediated spatial organization. This suggests a practical workflow for computational biology: use attention analysis for broad exploration of what patterns the model has learned about genomic organization, then use gradient analysis to identify specific prediction-critical features that merit experimental follow-up. The approximately 10% overlap we observed across 100 validation samples confirms that these approaches provide largely independent information, and both are needed for comprehensive interpretation.

One important caveat concerns the interpretation of gradient magnitudes across modalities. In our analysis, DNA gradients were substantially larger than RNA or protein gradients, which might suggest that RNA and protein inputs contribute minimally to predictions. However, this asymmetry is an artifact of our experimental design rather than a biological conclusion. DNA input varies across samples (each enhancer has different sequence), while RNA and protein inputs are fixed (the same pre-trained embeddings for all samples). Since prediction differences between samples can only arise from DNA differences, DNA gradients must be large by construction. RNA and protein provide fixed contextual information that shapes how DNA features are interpreted, but their gradients are necessarily small in this setup. Future extensions with cell-specific RNA inputs—where each cell's expression profile differs—would produce larger RNA gradients, potentially enabling the tracing of cell-to-cell variation in enhancer effects to specific gene expression differences.

The methodological framework we have outlined—attention for exploration, gradients for identifying prediction-critical features—provides tools for discovery, but these tools have important limitations that must be acknowledged.

### 4.3 Limitations and Caveats

Several limitations constrain the conclusions of this work. All experiments used K562 cells from a single dataset (Gasperini et al.), and enhancer function is highly cell-type specific—learned patterns may not generalize to other cellular contexts. The prediction accuracy itself is moderate: Pearson $r = 0.503$ corresponds to $R^2 = 0.25$, meaning the model explains only 25% of variance in enhancer effects. The gap between training ($r = 0.65$) and validation ($r = 0.503$) performance indicates overfitting, suggesting that some learned patterns may be dataset-specific rather than generalizable. The mechanistic validation

(FNDC5/CTCF) represents a single case study; systematic validation across many enhancer-gene pairs would strengthen confidence in gradient analysis as a discovery tool.

This work presents CDT v1, where RNA and protein embeddings are fixed across all samples—specifically, we use static gene embeddings from scGPT's embedding layer rather than contextualized representations from its transformer. This design choice reflects our proof-of-concept approach: by using fixed gene token embeddings, we prioritize avoiding data leakage over capturing cell-specific expression patterns. These fixed embeddings capture general gene properties learned during scGPT's pre-training on 33 million cells, but do not reflect cell-type-specific expression patterns. Consequently, Self-Attention patterns on RNA and protein modalities do not vary across samples and cannot directly explain prediction differences. CDT v2 will address this limitation by incorporating cell-type-specific contextualized embeddings. By passing cell-type-specific expression profiles through scGPT's full transformer architecture, v2 will generate dynamic RNA representations that capture cellular context. This will enable: (1) cell-type-specific enhancer effect prediction, (2) meaningful Self-Attention analysis across RNA and protein modalities, and (3) full utilization of scGPT's learned representations beyond the static embedding layer.

Attention interpretation faces an additional challenge: head diversity. CDT's 8 attention heads learn markedly different patterns (Figure 7), ranging from concentrated patterns to diffuse global attention. This diversity means that different heads may highlight different genomic positions as "important," complicating biological interpretation. Averaging across heads, while common practice, may obscure patterns captured by individual heads. Furthermore, DNA→RNA cross-attention shows uniform patterns across all genes (correlation > 0.99) due to fixed RNA context and single-window DNA input centered only on enhancers. Gene TSS regions are not directly represented, limiting the model's ability to learn explicit enhancer-gene spatial relationships. A dual-window architecture incorporating both enhancer-centered and gene TSS-centered DNA embeddings would enable DNA-DNA cross-attention to capture these spatial relationships directly. Until then, gradient analysis provides a valuable complement: gradients are head-invariant, identifying prediction-critical positions regardless of how attention is distributed across heads or whether cross-attention captures gene-specific patterns.

We deliberately used enhancer-centered (not TSS-centered) DNA sequences to avoid data leakage—the DNA sequence window contains no explicit positional information about transcription start sites, preventing the model from exploiting TSS proximity as a shortcut. This design choice strengthens reliability but may sacrifice predictive power. The achievement of $r = 0.503$ under this stringent setting suggests genuine sequence features drive predictions, but incorporating TSS information in future work could improve performance while requiring careful controls for leakage. More broadly, the current implementation uses 114kb DNA windows (not whole genomes), fixed gene embeddings (not cell-specific profiles), and a single prediction task (not general cellular modeling). These constraints define the proof-of-concept nature of this work; full realization of the Central Dogma Transformer framework awaits advances in foundation models and multi-omics data.

The 114kb window limitation has specific implications for understanding chromatin architecture. CTCF-mediated loops typically involve pairs of CTCF binding sites in convergent orientation, with cohesin maintaining the loop structure between them. In the FNDC5 case study, gradient analysis identified one CTCF site at −56.7kb from the enhancer, but a second CTCF anchor would be expected for loop formation; this region (near FNDC5 at −726kb) lies far outside the observable window. This means CDT can only identify regulatory elements within ±57kb of the enhancer center, potentially missing the complete loop architecture. A model capable of analyzing whole-genome sequences could identify both CTCF anchors simultaneously, enabling full characterization of chromatin loop structures and their regulatory consequences. This represents an important direction for future development: extending

CDT to incorporate genomic context beyond the current window, either through hierarchical attention mechanisms or integration with chromosome-scale language models.

## 4.4 Experimental Feedback Loop

CDT's potential value as a scientific tool lies in generating testable hypotheses. Attention and gradient analyses identify specific genomic positions, genes, and proteins as potentially important; experiments can test these predictions; results feed back to improve the model. The workflow illustrated earlier—computational prediction followed by database queries for supporting evidence—represents only the first step. We emphasize that querying existing databases differs from prospective experimental validation; definitive confirmation of computationally-identified regulatory elements would require perturbation experiments. Failed predictions would be equally valuable, revealing model limitations and guiding improvements.

## 4.5 The Path Forward: A Multi-Decade Vision

CDT represents a first step, not a destination. The constraints outlined in the Limitations section—restricted genomic windows, fixed embeddings, a single prediction task—define the proof-of-concept nature of this work. Overcoming these constraints requires not only technical advances but also new data resources: whole-genome regulatory maps, cell-specific multi-omics profiles across diverse contexts, and systematic perturbation datasets spanning cell types and developmental stages.

**Architectural Extensions: Feedback Loops**

The current CDT architecture implements a unidirectional flow following the classical Central Dogma: DNA → RNA → Protein. However, biological systems include regulatory feedback where proteins influence gene expression—transcription factors bind to DNA to activate or repress transcription, and signaling proteins modify chromatin state. A natural architectural extension is to incorporate a Protein → DNA cross-attention layer, creating a feedback loop where protein representations can query DNA representations. This would enable modeling of how protein states influence transcriptional regulation, with attention maps revealing which proteins the model considers relevant for regulating each genomic region. Such feedback mechanisms present both opportunities (capturing regulatory dynamics) and challenges (training stability, interpretation complexity). The current work establishes the viability of the unidirectional architecture; investigating feedback extensions is a natural next step.

**Per-Cell RNA Embeddings**

The current architecture uses fixed scGPT gene token embeddings for RNA representations—a deliberate design choice for this proof-of-concept that prioritizes avoiding data leakage over capturing cell-specific expression patterns. This uniformity limits cross-attention interpretability but does not constrain predictive power, as sample-specific information flows through DNA embeddings. A natural extension is incorporating per-cell RNA embeddings derived from single-cell RNA sequencing data, where each sample would have RNA representations reflecting its specific expression state. This would enable cross-attention patterns that vary meaningfully across samples, potentially revealing how expression context influences regulatory relationships. Such extensions await appropriate datasets pairing CRISPRi perturbations with single-cell transcriptomics.

**Additional Data Modalities**

Future versions might incorporate additional data modalities such as Hi-C chromatin structure, epigenetic modifications, or AlphaFold protein structures—each presenting substantial integration challenges. Such advances require sustained collaboration across computational biology, experimental biology, and machine learning. The ultimate goal is AI systems that accurately model how living cells work: how genetic information flows through transcription and translation to produce phenotypes, how molecular components regulate this flow, and how perturbations propagate through the system. Such capability could transform medicine, enable more rational approaches to drug design, and reveal fundamental principles of biological organization. CDT is a small step toward this ambitious goal—an architectural framework embodying principles we believe essential for mechanism-oriented biological AI.

**Advanced Attention Interpretation Methods**

The current attention analysis averages across all attention heads, which may obscure head-specific patterns. Each head in a multi-head attention layer can learn distinct functions—some may focus on proximal regulatory elements while others capture distal interactions. Recent advances in attention interpretation offer promising alternatives: Attention Rollout tracks attribution paths by multiplying attention matrices across layers; Attention Flow treats attention as a flow network solved as a max-flow problem [37]; and GMAR (Gradient-Driven Multi-Head Attention Rollout) [38] weights heads by their task-specific gradients, prioritizing heads that contribute most to predictions. Applying these methods to CDT could reveal which attention heads specialize in different regulatory mechanisms and provide more nuanced interpretability than simple averaging. Investigating these advanced interpretation techniques represents a natural direction for enhancing CDT's mechanistic insights.

**Toward Mechanism-Oriented AI for Science**

The remarkable success of modern AI—from large language models to autonomous driving—has been achieved through what might be termed *task-oriented* approaches: systems optimized to maximize performance on specific tasks without regard for how their internal representations relate to the underlying structure of the domain. This approach has proven extraordinarily effective for engineering applications where prediction accuracy is the primary goal. However, science demands more than prediction. The goal of scientific research is to understand the mechanisms that produce phenomena—in biology, how molecular interactions give rise to cellular behavior. Task-oriented AI, however accurate, cannot provide this mechanistic insight because its internal representations bear no necessary relationship to the causal structure of the domain it models.

CDT represents an alternative approach that we term *mechanism-oriented*: the computational architecture reflects the structure of the domain, so that what the model learns becomes interpretable in domain-specific terms. When CDT's attention flows from DNA to RNA, it parallels transcriptional regulation; when attention flows from RNA to protein, it mirrors translation. This architectural alignment transforms attention weights from opaque computational artifacts into interpretable representations of biological relationships. The gradient analysis that identified a CTCF binding site as prediction-critical is not merely a computational sensitivity—it traces backward through the same causal chain that operates in biological gene regulation.

This distinction has implications beyond CDT. As AI systems become increasingly central to scientific discovery, the choice between task-oriented and mechanism-oriented architectures may determine whether AI accelerates scientific understanding or merely substitutes prediction for explanation. We suggest that mechanism-oriented design—where computational structure mirrors domain structure—offers a path toward AI systems that genuinely advance scientific knowledge rather than replacing the need for it.

# 5. CDT as a Collaborative Ecosystem

Current AI development in biology, as in other fields, largely follows scaling laws: larger models trained on more data with more compute tend to yield better performance. While this paradigm has driven remarkable progress, it raises concerns about equity in scientific research. Training state-of-the-art foundation models now requires computational resources—thousands of GPUs running for months—that are available only to a handful of well-funded institutions and industry partners. This concentration of capability risks creating a landscape where many research groups can neither develop nor adequately validate the AI systems that increasingly shape biological discovery. The scientific community's ability to scrutinize, reproduce, and build upon foundational work may be compromised if only a few organizations possess the resources to train and evaluate large-scale models.

We propose CDT as one possible approach to this challenge: an integrative ecosystem that builds upon the collective achievements of the global research community. Rather than training a monolithic model from scratch, CDT leverages specialized foundation models, each representing years of dedicated research by different groups worldwide. This design philosophy enables researchers with modest computational resources to participate in multi-modal biological AI, extending the benefits of foundation model research to the broader scientific community.

CDT was designed from the outset as a modular architecture with interchangeable components. As DNA foundation models advance—from Enformer to models like Evo 2 [7] with million-base-pair context windows—CDT can incorporate these improvements by simply replacing the DNA encoder. Similarly, expanded RNA models (including non-coding transcripts) could upgrade the RNA encoder, and improved protein models incorporating structure and function could enhance the protein encoder. The integration layers—projection layers that map each modality to a shared dimension, cross-attention layers that connect components, and the prediction head—are deliberately lightweight, containing far fewer parameters than the underlying foundation models. When a component is upgraded, only these integration layers require fine-tuning to adapt to the new embedding space—a computationally modest operation that enables the architecture to incorporate advances in the field without full retraining.

This modular design enables a powerful synergy between general-purpose and domain-specific models. Large-scale efforts by well-resourced institutions produce increasingly powerful general-purpose DNA language models; meanwhile, individual research groups can train specialized models on their own experimental data—immunology-specific single-cell atlases, cancer-specific protein expression profiles, or tissue-specific regulatory maps. CDT provides the framework to combine these complementary resources: an Immunology-CDT might pair Evo 2′s general DNA understanding with immune cell-specific RNA embeddings and antibody protein representations; a Cancer-CDT could integrate tumor-specific expression models; a Neuroscience-CDT could incorporate brain cell transcriptome data. Each variant benefits from both the general knowledge encoded in large foundation models and the specialized knowledge derived from domain-specific experiments.

This ecosystem approach means that everyone can contribute and everyone benefits. Different research groups might develop specialized variants for their biological contexts, each adapting the framework to domain-specific data while sharing architectural innovations. Attention and gradient maps could provide a common interpretive language across variants: convergent predictions across contexts would strengthen confidence in identified mechanisms, while divergent predictions would highlight context-specific regulation worth investigating. Realizing this vision would require open-sourcing the architecture, pre-trained weights, and analysis pipelines to lower barriers for the research community—a direction we intend to pursue as the framework matures.

# 6. Conclusion

The Central Dogma Transformer demonstrates that aligning AI architecture with biological information flow enables both accurate prediction and mechanistic interpretation. This proof-of-concept achieves a Pearson correlation of 0.503 for enhancer effect prediction—63% of the theoretical ceiling imposed by experimental variability—while providing interpretable attention and gradient maps that highlight regulatory features.

Three key findings emerge from this work. First, attention and gradient analyses capture complementary aspects of model behavior, with largely distinct genomic regions highlighted in detailed case studies —motivating a practical "Two Windows" workflow for biological discovery. Second, gradient analysis identified a CTCF binding site positioned within a region where Hi-C data showed physical contact between the enhancer and target gene, suggesting the model learned features relevant to chromatin architecture. Third, the modular design enables straightforward integration of improved foundation models as they become available.

The current work has clear limitations: a restricted 114kb genomic window, fixed pre-computed embeddings, and validation in a single cell type. These constraints define CDT v1 as a proof of concept. Nevertheless, the results indicate that mechanism-oriented architectures—where computational structure reflects domain structure—offer a viable path toward interpretable biological AI. Unlike task-oriented approaches that optimize prediction without regard for interpretability, CDT demonstrates that accuracy and mechanistic insight need not be in tension. This work takes a step toward AI systems that reveal, not merely predict, the regulatory logic of living cells.

---

# 7. Methods

## 7.1 Data Sources

**CRISPRi Screen Data**: We used the Gasperini et al. (2019) [29] at-scale CRISPRi enhancer screen (GEO accession GSE120861), which systematically targeted 5,920 candidate enhancer elements in K562 cells and measured effects on gene expression via single-cell RNA-seq (207,324 cells × 13,135 genes). Candidate enhancers were selected based on DNase I hypersensitivity sites (DHS)—regions of open chromatin that indicate potential regulatory activity—excluding promoter-proximal regions near transcription start sites. Each enhancer-gene pair has an associated effect size (beta value) quantifying the change in gene expression upon enhancer perturbation.

**Training Labels**: Beta values from the CRISPRi screen serve as regression targets. Negative beta values indicate activating enhancers (perturbation reduces expression); positive values indicate repressive elements.

**Cross-Experiment Reference Data**: For gene identifier mapping and cross-experiment validation, we used the Morris et al. STING-seq dataset [30], an independent CRISPRi screen in K562 cells that provides ENSG-to-gene-symbol mappings and serves as a reference for reproducible K562 gene expression.

## 7.2 Foundation Model Embeddings

All foundation models are completely frozen, with embeddings pre-computed once and cached to disk, enabling efficient training without fine-tuning the underlying foundation models.

**DNA Embeddings (Enformer)**: For each enhancer in the Gasperini dataset, we computed the enhancer center as the midpoint of the annotated enhancer coordinates (start + end) / 2. We then extracted 196,608 bp of DNA sequence centered on this midpoint from the hg38 reference genome. Enformer [3] processes this input sequence and outputs embeddings for the central 114,688 bp region, represented as 896 positional bins (128 bp per bin). We extracted embeddings from Enformer's trunk layer (before the prediction heads), obtaining 3,072-dimensional features per position. This yields 896 × 3,072 dimensional embeddings per enhancer, capturing long-range regulatory sequence features within ±57 kb of each enhancer center.

**RNA Embeddings (scGPT)**: scGPT [8] represents genes as tokens in a vocabulary, with gene tokens and expression values embedded separately. We extracted static (data-independent) gene token embeddings by passing each gene's vocabulary ID through scGPT's encoder layer, obtaining 512-dimensional embeddings without expression value input. This approach captures general gene properties learned during scGPT's pre-training on 33 million human cells, independent of cell-type-specific expression contexts. Genes were matched by official gene symbol.

**Protein Embeddings (ProteomeLM)**: We downloaded the human proteome from UniProt [39] (Swiss-Prot reviewed entries, organism ID 9606) in FASTA format, extracting gene names from the GN= field in sequence descriptions. Protein sequences were processed using the ProteomeLM pipeline [12]: ESM-C first computed sequence embeddings for each protein. Simultaneously, ortholog group embeddings were constructed by averaging ESM-C embeddings across proteins belonging to the same orthologous group (annotations from OrthoDB [40]). Both the sequence embeddings and ortholog group embeddings were then processed through ProteomeLM (112M parameters), a transformer trained on 32,000 proteomes spanning all domains of life. We extracted the final hidden state (768 dimensions) as protein embeddings. Genes without UniProt entries were excluded from the final gene set.

### 7.3 Gene Selection and Alignment

Gene selection was based on the intersection of three criteria: (1) genes detected in both the Gasperini et al. CRISPRi scRNA-seq data (GSE120861) and the Morris et al. STING-seq dataset (ensuring reproducible K562 expression across independent experiments), (2) genes present in scGPT's vocabulary, and (3) genes with available protein embeddings in ProteomeLM. This intersection yielded 2,360 genes for model training and evaluation.

The same 2,360-gene set is used for both RNA and protein embeddings, ensuring alignment across modalities. Training data from the Gasperini CRISPRi screen confirms these genes are present in both experiments, as beta values can only be calculated for genes with detectable expression. By requiring expression in both datasets, we select genes with reproducible K562 expression, reducing the influence of technical artifacts on our analysis.

### 7.4 Cross-Modal Alignment

CDT integrates three modalities with fundamentally different data structures. DNA embeddings are sample-specific: each enhancer has unique embeddings derived from its genomic context (896 positions × 3,072 dimensions per sample). In contrast, RNA and protein embeddings are shared across samples: the same 2,360-gene embedding matrices are used for all training examples, representing static gene and protein properties independent of the specific enhancer being analyzed.

This asymmetry reflects the biological task: CDT predicts how perturbing a specific enhancer (variable DNA context) affects expression of genes (fixed gene set). The cross-attention mechanism allows each sample's DNA embeddings to query the shared RNA embeddings, enabling the model to learn which genomic features associate with which genes' regulation.

**RNA-Protein Alignment.** For the RNA→Protein cross-attention to be meaningful, position $i$ in the RNA embedding tensor must correspond to the same gene as position $i$ in the protein tensor. We adopted HGNC gene symbols as the common alignment key:

1. **Gasperini CRISPRi data**: The original dataset uses ENSG (Ensembl) identifiers. We converted these to gene symbols using the Morris STING-seq dataset's gene annotation. This conversion also serves as a filter—only genes present in both independent K562 experiments are retained.

2. **scGPT embeddings**: The vocabulary is symbol-based, enabling direct lookup. We extracted embeddings only for the 2,360 aligned genes.

3. **ProteomeLM embeddings**: We extracted gene symbols from UniProt's GN= field and generated embeddings only for proteins in the aligned gene set.

**Canonical Ordering.** The aligned gene set was sorted alphabetically to ensure reproducibility. RNA and protein embedding files store genes in this canonical order, guaranteeing that index $i$ refers to the same gene in both modalities ($n_{\text{genes}} = n_{\text{proteins}} = 2,360$).

**Index Mapping.** Training data references genes by their original dataset indices. We pre-computed a mapping from original indices to aligned indices, enabling the data loader to retrieve correct embeddings without regenerating training files.

**Verification.** Alignment correctness was verified by confirming that 100 randomly sampled indices return identical gene symbols across RNA and protein embedding files. This verification executes automatically during model initialization.

### 7.5 Training Data Preparation

From the Gasperini CRISPRi screen, we extracted enhancer-gene pairs with effect sizes and applied the gene alignment filter, resulting in 4,605 training samples and 996 validation samples. Data was split at the enhancer level (not the pair level) to prevent data leakage—all pairs involving a given enhancer appear in either training or validation, never both.

Each training sample contains: (1) DNA embedding index (mapping to pre-computed Enformer embeddings), (2) target gene index (for extracting the relevant prediction from CDT's output), and (3) beta value (regression target).

### 7.6 Model Architecture

CDT implements a multi-modal transformer architecture [2] illustrated in Figure 1. The model processes three input modalities through a series of attention-based layers, following the information flow of the Central Dogma.

**Architecture Overview.** The forward pass proceeds through six stages:

1. **Input Layer**: Pre-computed embeddings from three foundation models are loaded—DNA from Enformer (shape: 896 × 3,072), RNA from scGPT (shape: $n_{\text{genes}} \times 512$), and Protein from ProteomeLM (shape: $n_{\text{genes}} \times 768$), where $n_{\text{genes}} = 2,360$.

2. **Projection Layer**: Linear projections with layer normalization and dropout map each modality into a shared 768-dimensional space, enabling cross-modal attention operations.

3. **Self-Attention Layers**: Each modality undergoes independent self-attention to refine representations before cross-modal integration. DNA receives 2 layers (capturing long-range genomic interactions), while RNA and Protein each receive 1 layer (capturing co-expression and protein-protein relationships).

4. **Cross-Attention Layers**: Information flows unidirectionally following the Central Dogma. First, RNA queries DNA (modeling transcriptional regulation), producing RNA representations fused with genomic context. Then, Protein queries the fused RNA (modeling translational relationships), producing Protein representations that incorporate both RNA and DNA information.

5. **Virtual Cell Embedder (VCE)**: Attention-based pooling integrates all three modalities into a single 768-dimensional cell-state vector. Learned query vectors attend over each modality, and the pooled representations are concatenated and passed through a feed-forward network.

6. **Task Head**: A linear layer maps the VCE output to predictions for all $n_{\text{genes}}$ genes simultaneously.

**Model Configuration.** The hyperparameters used in this study are:

| Component | Configuration |
| --- | --- |
| Hidden dimension ($d$) | 768 |
| Attention heads ($h$) | 8 |
| Head dimension ($d_k = \frac{d}{h}$) | 96 |
| DNA Self-Attention layers | 2 |
| RNA Self-Attention layers | 1 |
| Protein Self-Attention layers | 1 |
| Cross-Attention layers | 2 (DNA→RNA, RNA→Protein) |
| Dropout probability | 0.3 |
| Total trainable parameters | 60M |

Table 3: **CDT Model Configuration.**

**Training Output.** The model outputs predictions for all genes simultaneously as a vector $\hat{\boldsymbol{y}} \in \mathbb{R}^{n_{\text{genes}}}$. During training, only the prediction corresponding to the target gene of each sample contributes to the loss—other predictions are masked. This design enables efficient batched training while maintaining per-gene prediction capability at inference time.

### 7.7 Projection Layer Formulation

Each modality's embeddings are projected into a shared $d$-dimensional space ($d = 768$) using learned linear projections with layer normalization and dropout:

$$\boldsymbol{H}_m = \text{Dropout}(\text{LayerNorm}(\boldsymbol{X}_m \boldsymbol{W}_m + \boldsymbol{b}_m))$$

where $\boldsymbol{X}_m$ is the input embedding matrix for modality $m$, and $\boldsymbol{W}_m \in \mathbb{R}^{d_m \times d}$ is the projection matrix. The input dimensions $d_m$ vary by modality:

- DNA (Enformer): $d_{\text{DNA}} = 3072 \rightarrow 768$
- RNA (scGPT): $d_{\text{RNA}} = 512 \rightarrow 768$
- Protein (ProteomeLM): $d_{\text{Protein}} = 768 \rightarrow 768$

This projection enables cross-modal interactions by placing all representations in a common dimensional space, while layer normalization stabilizes training and dropout ($p = 0.3$) provides regularization.

## 7.8 Cross-Attention Formulation

CDT employs directional cross-attention to implement the Central Dogma information flow. The standard scaled dot-product attention is defined as:

$$\text{Attention}(\boldsymbol{Q}, \boldsymbol{K}, \boldsymbol{V}) = \text{softmax}\left(\frac{\boldsymbol{Q}\boldsymbol{K}^T}{\sqrt{d_k}}\right)\boldsymbol{V}$$

where $\boldsymbol{Q} \in \mathbb{R}^{n_q \times d}$, $\boldsymbol{K} \in \mathbb{R}^{n_k \times d}$, and $\boldsymbol{V} \in \mathbb{R}^{n_k \times d}$ are the query, key, and value matrices respectively, and $d_k = \frac{d}{h}$ is the dimension per attention head (with $h = 8$ heads).

**DNA → RNA Cross-Attention (Transcription)**: RNA representations query DNA representations to gather regulatory information:

$$\boldsymbol{H}_{\text{RNA}}^{\text{fused}} = \boldsymbol{H}_{\text{RNA}} + \text{CrossAttention}(\boldsymbol{Q} = \boldsymbol{H}_{\text{RNA}}, \boldsymbol{K} = \boldsymbol{H}_{\text{DNA}}, \boldsymbol{V} = \boldsymbol{H}_{\text{DNA}})$$

where $\boldsymbol{H}_{\text{RNA}} \in \mathbb{R}^{n_{\text{genes}} \times d}$ and $\boldsymbol{H}_{\text{DNA}} \in \mathbb{R}^{896 \times d}$. The attention weights $\boldsymbol{A}_{\text{DNA} \to \text{RNA}} \in \mathbb{R}^{n_{\text{genes}} \times 896}$ reveal which genomic positions each gene attends to—interpretable as potential regulatory associations.

**RNA → Protein Cross-Attention (Translation)**: Protein representations query the RNA-fused representations:

$$\boldsymbol{H}_{\text{Protein}}^{\text{fused}} = \boldsymbol{H}_{\text{Protein}} + \text{CrossAttention}(\boldsymbol{Q} = \boldsymbol{H}_{\text{Protein}}, \boldsymbol{K} = \boldsymbol{H}_{\text{RNA}}^{\text{fused}}, \boldsymbol{V} = \boldsymbol{H}_{\text{RNA}}^{\text{fused}})$$

where $\boldsymbol{H}_{\text{Protein}} \in \mathbb{R}^{n_{\text{genes}} \times d}$ (since $n_{\text{proteins}} = n_{\text{genes}}$ in our aligned gene set). The attention weights $\boldsymbol{A}_{\text{RNA} \to \text{Protein}} \in \mathbb{R}^{n_{\text{genes}} \times n_{\text{genes}}}$ capture transcript-protein associations.

This sequential structure ensures unidirectional information flow: DNA information reaches Protein only through RNA, mirroring the biological Central Dogma.

## 7.9 Virtual Cell Embedder (VCE) Formulation

The VCE integrates three modalities using attention-based pooling. For each modality $m \in \{\text{DNA}, \text{RNA}, \text{Protein}\}$, a learned query vector $\boldsymbol{q}_m \in \mathbb{R}^d$ attends over the sequence of representations $\boldsymbol{H}_m \in \mathbb{R}^{n_m \times d}$:

$$\boldsymbol{z}_m = \text{Attention}(\boldsymbol{q}_m, \boldsymbol{H}_m, \boldsymbol{H}_m) = \text{softmax}\left(\frac{\boldsymbol{q}_m \boldsymbol{H}_m^T}{\sqrt{d}}\right)\boldsymbol{H}_m$$

where $n_m$ is the sequence length for modality $m$ (896 for DNA, $n_{\text{genes}}$ for both RNA and Protein) and $d = 768$. The attention weights $\boldsymbol{a}_m = \text{softmax}\left(\frac{\boldsymbol{q}_m \boldsymbol{H}_m^T}{\sqrt{d}}\right) \in \mathbb{R}^{n_m}$ are interpretable, revealing which positions, genes, or proteins the model considers most relevant.

The pooled representations are concatenated and fused through a feed-forward network:

$$\boldsymbol{h}^{\text{VCE}} = \text{FFN}([\boldsymbol{z}_{\text{DNA}}; \boldsymbol{z}_{\text{RNA}}; \boldsymbol{z}_{\text{Protein}}])$$

where $\text{FFN}: \mathbb{R}^{3d} \to \mathbb{R}^d$ consists of two linear layers with GELU activation:

$$\text{FFN}(\boldsymbol{x}) = \boldsymbol{W}_2 \cdot \text{GELU}(\boldsymbol{W}_1 \boldsymbol{x} + \boldsymbol{b}_1) + \boldsymbol{b}_2$$

The final VCE output $\boldsymbol{h}^{\text{VCE}} \in \mathbb{R}^d$ represents the integrated cellular state.

## 7.10 Training Configuration

| Parameter | Value |
| --- | --- |
| Optimizer | AdamW [41] |
| Learning rate | $1 \times 10^{-4}$ |
| Weight decay | $1 \times 10^{-5}$ |
| Loss function | Huber ($\delta = 1.0$) |
| Batch size | 8 |
| Epochs | 50 (early stopping) |
| Patience | 10 epochs |
| Scheduler | ReduceLROnPlateau |

Table 4: **Training Hyperparameters.**

Training was performed on a single NVIDIA A100 GPU. The model achieved best validation performance at epoch 16 with Pearson r = 0.503.

## 7.11 Loss Function

The model is trained to minimize the Huber loss between predicted effect sizes $\hat{\beta}_i$ and ground truth beta values $\beta_i$ from the Gasperini et al. dataset:

$$L_{\text{Huber}}(\beta, \hat{\beta}) = \frac{1}{n} \sum_{i=1}^{n} \ell_\delta(\beta_i - \hat{\beta}_i)$$

where the element-wise loss $\ell_\delta$ is defined as:

$$\ell_\delta(r) = \begin{cases} \frac{1}{2} r^2 & \text{if } |r| \leq \delta \\ \delta(|r| - \frac{1}{2}\delta) & \text{otherwise} \end{cases}$$

with $\delta = 1.0$. The Huber loss combines mean squared error (MSE) for small residuals with mean absolute error (MAE) for large residuals, providing robustness to outliers in effect size measurements.

## 7.12 Attention and Gradient Analysis

**Attention Map Extraction**: Cross-attention weights are extracted during the forward pass by retaining intermediate attention matrices. The DNA→RNA attention map (shape: batch × heads × $n_{\text{genes}}$ × 896) shows which DNA positions each gene attends to during transcription-like information flow. For visualization, attention weights are averaged across the 8 attention heads.

**Gradient Analysis**: For a given prediction $\hat{\beta}$, we compute the gradient with respect to each DNA position's embedding:

$$g_p = \frac{\partial \hat{\beta}}{\partial x_p^{\text{DNA}}}$$

where $x_p^{\text{DNA}} \in \mathbb{R}^d$ is the $d$-dimensional embedding at genomic position $p$. The gradient importance score is the L2 norm:

$$I_p^{\text{gradient}} = \| g_p \|_2 = \sqrt{\sum_{j=1}^{d} \left( \frac{\partial \hat{\beta}}{\partial x_{p,j}^{\text{DNA}}} \right)^2}$$

By the chain rule, gradients flow through the CDT architecture in reverse:

$$\frac{\partial \hat{\beta}}{\partial x_p^{\text{DNA}}} = \frac{\partial \hat{\beta}}{\partial h^{\text{VCE}}} \cdot \frac{\partial h^{\text{VCE}}}{\partial h^{\text{Protein}}} \cdot \frac{\partial h^{\text{Protein}}}{\partial h^{\text{RNA}}} \cdot \frac{\partial h^{\text{RNA}}}{\partial x_p^{\text{DNA}}}$$

This "reverse Central Dogma tracing" attributes predictions to DNA positions through a pathway mirroring the reverse of biological information flow.

### 7.13 External Validation

**Hi-C Data**: Chromatin contact maps for K562 cells were obtained from the 4D Nucleome Data Portal (4DN). We used Hi-C data at 5kb resolution to examine chromatin contacts between genomic regions identified by attention and gradient analyses. Contact frequency between two genomic bins is quantified as the observed/expected ratio, where higher values indicate stronger physical proximity in 3D nuclear space.

For validation of CDT findings, we examined whether genomic positions highlighted by attention or gradient analysis showed elevated contact frequency with the target gene's promoter region. This provides orthogonal evidence for regulatory relationships: if CDT identifies a position as important for a gene's regulation, physical contact between that position and the gene (as measured by Hi-C) supports the biological plausibility of the computational finding.

**Evaluation Metrics**: Model performance was evaluated using Pearson correlation coefficient between predicted and observed effect sizes:

$$r = \frac{\sum_{i=1}^{n} \left( \hat{\beta}_i - \overline{\hat{\beta}} \right)\left( \beta_i - \overline{\beta} \right)}{\sqrt{\sum_{i=1}^{n} \left( \hat{\beta}_i - \overline{\hat{\beta}} \right)^2} \sqrt{\sum_{i=1}^{n} \left( \beta_i - \overline{\beta} \right)^2}}$$

where $\hat{\beta}_i$ and $\beta_i$ are the predicted and observed effect sizes for sample $i$, and $\overline{\hat{\beta}}$ and $\overline{\beta}$ are their respective means. Pearson correlation captures the linear relationship between predictions and ground truth, with $r = 1$ indicating perfect positive correlation.

## 8. Acknowledgements

We thank Vaswani et al. for the Transformer architecture that provided the conceptual foundation for CDT's cross-attention design. We are grateful to the teams behind the foundation models used in this work: Enformer (Avsec et al.) for DNA sequence embeddings, scGPT (Cui et al.) for single-cell gene expression representations, and ESM-C/ProteomeLM for protein embeddings. We thank Gasperini et al. for their CRISPRi enhancer screen data essential for training and validation, and the 4D Nucleome Consortium for Hi-C datasets that enabled independent validation. We acknowledge the Chainer team at Preferred Networks for pioneering Define-by-Run automatic differentiation, and the PyTorch developers for the framework that made this work technically feasible.

We are grateful to David Agard (UCSF/Chan Zuckerberg Initiative), Shin-ichi Maeda and Daisuke Okanohara (Preferred Networks) for their insightful feedback on the manuscript.

Finally, we thank Google Colab for providing accessible GPU computing resources. This entire research was conducted using a single GPU through Colab's free and low-cost tiers, demonstrating that meaningful deep learning research in computational biology is now accessible to independent researchers without institutional computing infrastructure.

The manuscript was reviewed and edited with assistance from Claude (Anthropic). The author takes full responsibility for the scientific content and conclusions.

## 9. Code and Data Availability

**Code:** CDT source code is available at https://github.com/nobusama/CDT under the MIT License.

**Data:** Pre-computed embeddings (DNA, RNA, and Protein) and training data are available on Hugging Face at https://huggingface.co/datasets/nobusama17/cdt-embeddings.

All materials necessary to reproduce the experiments and analyses presented in this work are freely available without restrictions.